\documentclass[runningheads]{llncs}
\usepackage[T1]{fontenc}
\usepackage{graphicx}
\usepackage{booktabs}
\usepackage[accsupp]{axessibility}  
\usepackage{algorithm2e}
\RestyleAlgo{ruled}
\usepackage{multirow}
\usepackage{subcaption}
\usepackage{soul}
\usepackage[dvipsnames]{xcolor}
\usepackage{colortbl}
\usepackage[pagebackref,breaklinks,colorlinks,citecolor=accvblue]{hyperref}
\hypersetup{
    colorlinks=true,
}

\newcommand{\trajdif}{\texttt{TrajDiffuse}\xspace}

\definecolor{accvblue}{rgb}{0.12,0.49,0.85}

\begin{document}
\title{TrajDiffuse: A Conditional Diffusion Model for Environment-Aware Trajectory Prediction}
\titlerunning{Conditional Diffusion for Environment-Aware HTP}
%
\author{Qingze (Tony) Liu\inst{1} \and
Danrui Li\inst{1} \and
Samuel S. Sohn\inst{1} \and \\
Sejong Yoon\inst{2} \and
Mubbasir Kapadia \inst{1} \and
Vladimir Pavlovic\inst{1} 
}
\authorrunning{Q.T. Liu et al.}
%

\institute{Rutgers University, Piscataway, USA \\
\email{\{tony.liu; danrui.li; mubbasir.kapadia\}@rutgers.edu; \\
\{sss286; vladimir\}@cs.rutgers.edu} \\
\and The College of New Jersey, Ewing, USA \\
\email{yoons@tcnj.edu}}
\maketitle              
\begin{abstract}
Accurate prediction of human or vehicle trajectories with good diversity that captures their stochastic nature is an essential task for many applications. However, many trajectory prediction models produce unreasonable trajectory samples that focus on improving diversity or accuracy while neglecting other key requirements, such as collision avoidance with the surrounding environment. In this work, we propose \trajdif, a planning-based trajectory prediction method using a novel guided conditional diffusion model. We form the trajectory prediction problem as a denoising impaint task and design a map-based guidance term for the diffusion process. \trajdif is able to generate trajectory predictions that match or exceed the accuracy and diversity of the SOTA, while adhering almost perfectly to environmental constraints. We demonstrate the utility of our model through experiments on the nuScenes and PFSD datasets and provide an extensive benchmark analysis against the SOTA methods.

\end{abstract}

\begin{figure}
\vspace{-0.2in}
    \centering
        \includegraphics[width=0.85\linewidth]{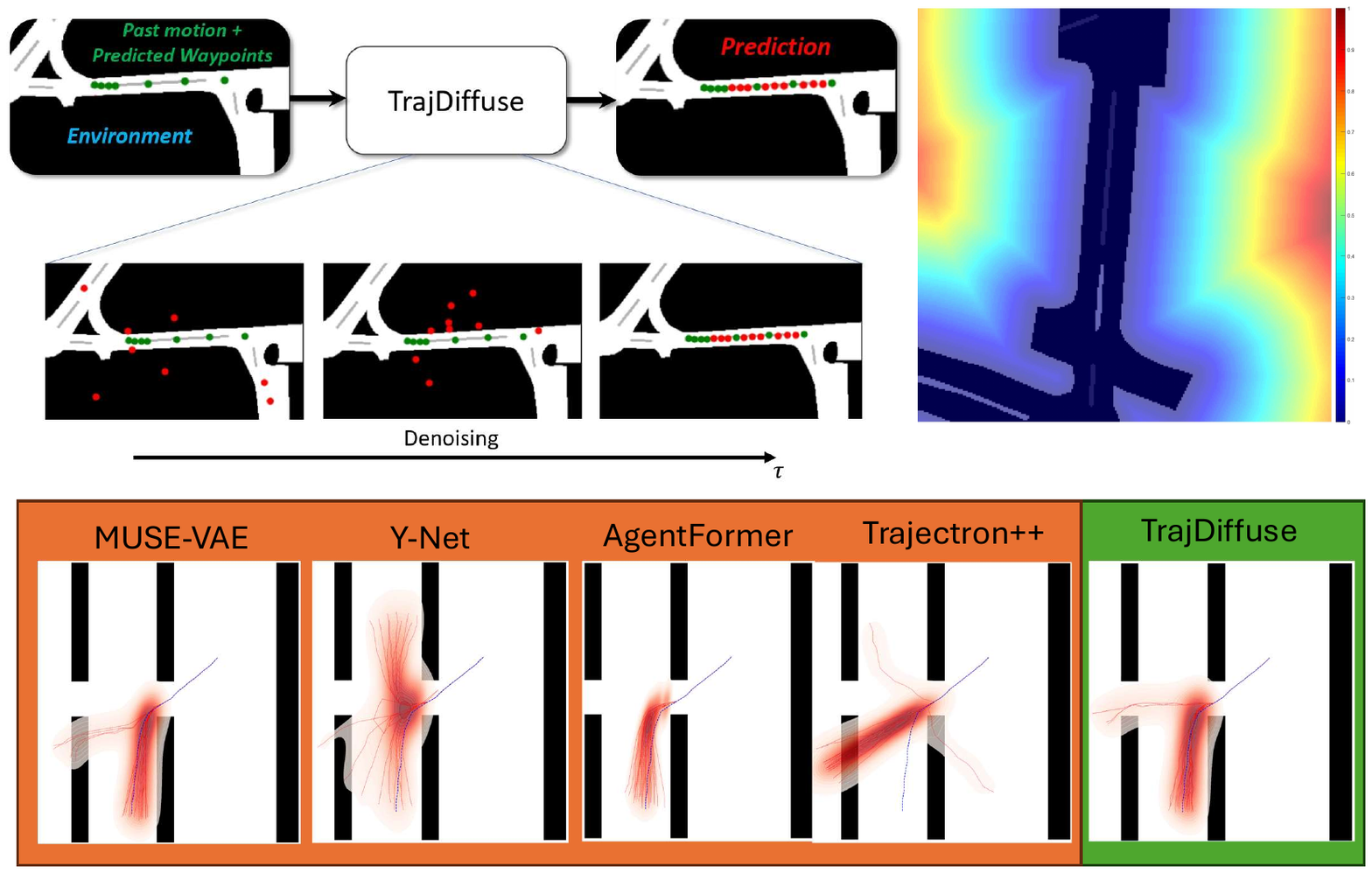}
        \label{fig:approach}
    \vspace{-0.2in}
    \caption{\textbf{Top Left}: Illustration of the denoising trajectory prediction process. {\color{Green} Green} dots indicate the observed trajectory and the predicted way points. {\color{Red} Red} dots are the denoised prediction. \textbf{Top Right}: A map from nuScene dataset overlayed with the distance transform showing the distance to the navigable areas (roads) in the scene. The color scale represents the normalized distance from closest (blue) to the farthest (red). \textbf{Bottom}: Comparison against other SOTA methods on PFSD dataset}
    \label{fig:fig1}
\vspace{-0.2in}
\end{figure}

\section{Introduction}
Recent Human Trajectory Prediction (HTP) \cite{alahi2016social,chai2019multipath,ivanovic2019trajectron,lee2022muse,mangalam2021goals,yuan2021agentformer} works have achieved great prediction accuracy by incorporating the stochastic nature of human trajectory; this means that HTP models sample multiple trajectory predictions and use the most accurate one for accuracy evaluation. HTP models also promote sample diversity to cover as many modes of movement as possible. However, SOTA models often sacrifice the feasibility and realism of the prediction in order to achieve the above-mentioned properties, failing to produce quality trajectory samples that consistently follow scene contexts. In this paper, we investigate how to achieve accurate and diverse predictions while making sure that the generated trajectories are also reasonable with respect to environmental constraints.

To predict an accurate multimodal distribution of human trajectories, recent work has often adopted a planning-based approach by learning a probabilistic latent variable model of the agent's intent of movement \cite{yuan2021agentformer,pang2021trajectory}. Such formulation models the human decision-making process in which human agents often have a goal in mind and execute their movement based on their goals. These methods typically use a CVAE-based model to decode a latent variable to obtain the trajectory prediction. Despite the effectiveness of this strategy, we argue that these works often lack explicit control over the generation and decoding process. 

Other non-probabilistic methods often rely on test-time sampling tricks or the anchoring approach \cite{mangalam2021goals,chai2019multipath}. These methods can achieve diverse sampling by using handcrafted criteria that promote the output trajectories to be distinct from each other. However, we have observed that the sampling trick will cause a trade-off between the diversity and the feasibility of predicted trajectories.

To address these issues, we propose a trajectory prediction method based on a guided conditional diffusion model. Compare with previous diffusion-based approaches \cite{Gu_2022_CVPR,mao2023leapfrog,li2023multi}, which use latent observation embedding as input, our method formulates the trajectory prediction problem as an interpolation between the agent trajectory history and the predicted agent motion intent, as illustrated on the top left of \autoref{fig:fig1}. By adopting the diffusion model framework, we gain the ability to directly and explicitly control the trajectory generation process via goal/waypoint conditioning and a guidance function that fuses the higher-level signal of an agent's intent with constraints such as the environmental awareness encoded on the top right of \autoref{fig:fig1}. Our method has shown better environmental understanding compared to other SOTA HTP methods as shown in both the bottom of \autoref{fig:fig1} and experiments in the later sections.

Our contributions are summarized as follows: (1) We propose a novel planning-based trajectory prediction algorithm \trajdif using a guided conditional diffusion model to obtain accurate and diverse trajectory predictions. (2) We propose an environment-based gradient guidance term that ensures that output trajectories are feasible and environment-compliant. (3) We demonstrate our design through experiments on two public datasets. We show that our method can produce an accurate trajectory prediction and achieve good compliance with environmental constraints. The source code can be found here: \url{https://github.com/TL-QZ/TrajDiffuse.git}

\section{Related Work}
\label{sec:related}

\paragraph{\textbf{Trajectory Prediction.}} Early works in trajectory prediction have focused on performing the trajectories prediction with sequence-to-sequence models such as the RNN-based Social-LSTM proposed by \cite{alahi2016social}. To capture the multimodality of human trajectories, recent work has often applied probabilistic generative frameworks for the prediction process. Social-GAN and Social-BiGAT\cite{gupta2018social,kosaraju2019social} proposed to use the Generative Adverseral Network (GAN) \cite{goodfellow2014generative} to generate multiple prediction outputs by repeatedly sampling inputs for the generative network. MUSE-VAE, Trajectron++ and Agentformer\cite{lee2022muse,salzmann2020trajectron++,yuan2021agentformer} applied the CVAE \cite{sohn2015learning} for inference on the distribution of latent agent intents. For nonprobabilistic approaches, Y-Net\cite{mangalam2021goals} used a test-time sampling trick to achieve diverse trajectory predictions. MultiPath\cite{chai2019multipath} leveraged a fixed set of state sequence anchors to generate diverse modes of trajectory predictions. GOHOME \cite{10.1109/ICRA46639.2022.9812253} uses lane information from HD-maps to generate heatmaps for intent prediction. Despite its effectiveness, such information are only viable for vehicle trajectory prediction, as pedestrian does not follows specific lanes. Besides, the evaluations of above-mentioned models often ignored feasibility and only promoted diversity and accuracy of the mode with least displacement from the ground truth, making generation unusable and unrealistic trajectory predictions. We have shown in our experiments that our model alleviates this problem by considering the environmental feasibility of each output prediction and guiding the sampling process with a map-based gradient correction term.

\paragraph{\textbf{Diffusion Models.}}
The recent success of diffusion models~\cite{ho2020denoising,sohl2015deep,song2019generative} in image and audio generation has shown their ability to generate high-quality, high-dimensional samples. The applications of diffusion models have also been extended to multimodal learning \cite{nichol2021glide}, sequence learning \cite{rasul2021autoregressive}, and offline reinforcement learning tasks~\cite{wang2022diffusion,janner2022diffuser}. Recent works also attempt to apply diffusion for the HTP task. Gu et al.~\cite{Gu_2022_CVPR} proposed Motion Indeterminacy Diffusion (MID) that predicts the agent's trajectory by denoising diffusion of Gaussian noise to trajectory predictions using the observed trajectory embeddings as an additional condition to the model. This approach requires a large number of diffusion steps.  Li et al.~\cite{li2023multi} and Mao et al.~\cite{mao2023leapfrog} proposed using a trajectory proposal rather than pure Gaussian noises as the start of the denoising process to reduce the number of steps required. Leapfrog Diffusion~\cite{mao2023leapfrog} generates the trajectory proposal from a learned deterministic trajectory initializer, while~\cite{li2023multi} used a CVAE-based module for this initialization. MotionDiffuser~\cite{Jiang_2023_CVPR} focused on learning the joint distribution for motions of multiple agents to generate trajectory predictions without agent-agent collision. The above-mentioned diffusion-based HTP models all uses latent embedding of observed trajectories as condition and generate trajectories via denoising diffusion from either Gaussian noises or a full trajectory proposal. We argue that such a setting lacks explicit control over the generation process and does not offer a clear strategy that incorporates environmental information. Inspired by recent work \cite{janner2022diffuser} on the use of the diffusion model for goal-conditioned offline reinforcement learning, we propose a new planning-based HTP model using guided conditional denoising diffusion to solve human trajectory prediction as an impainting task by interpolating the agent's trajectory history and predicted agent motion intent. We also propose a map-based gradient guidance term to ensure that all generated trajectory samples are complying with the environmental constraint. We compare with the MID model and the results of our experiments demonstrate the effectiveness of our model in both predictive accuracy and quality performance. For other diffusion-based models, we were not able to included in the benchmark due to the source code being unavailable .
 \section{Proposed Method}
\label{sec:method}
\begin{figure*}
    \centering
    \includegraphics[width=0.75\textwidth]{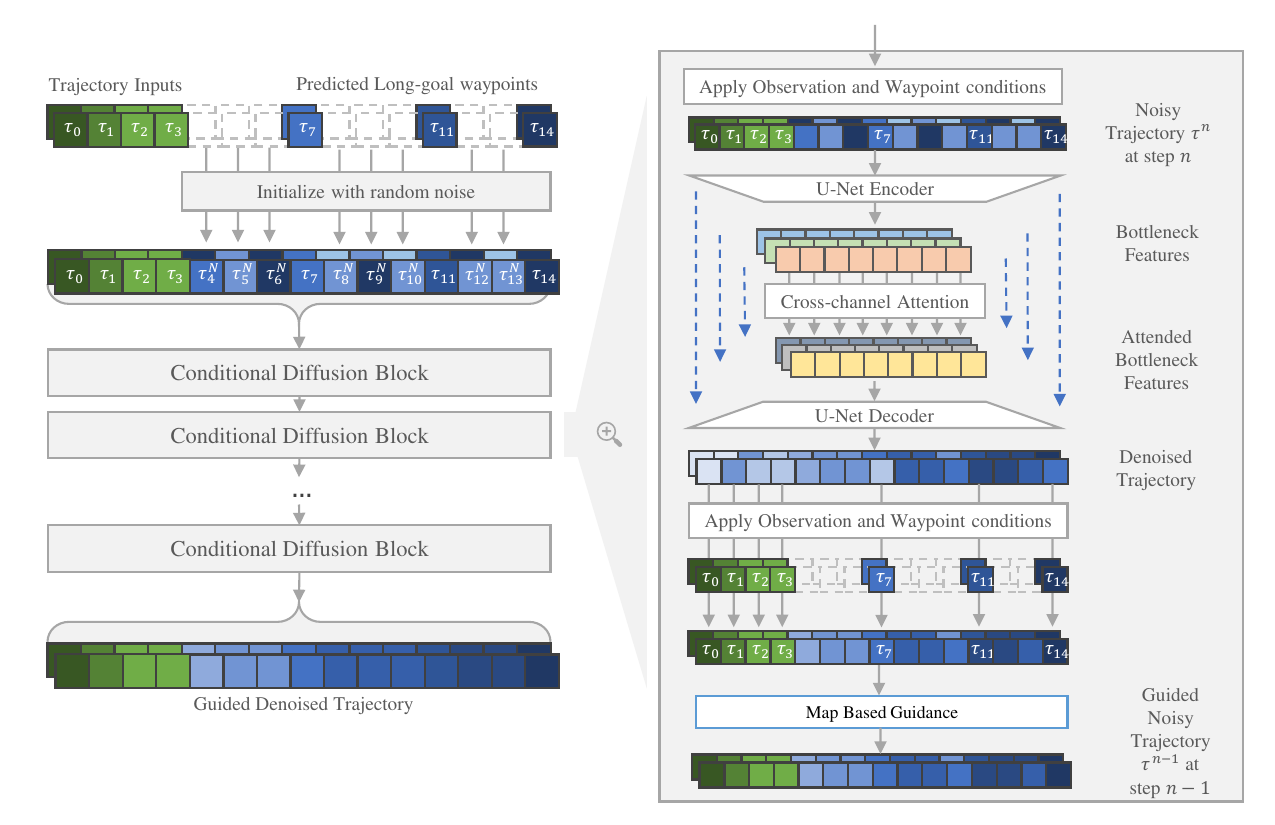}
     \caption{Details on \trajdif Model Structure.  Left: Prediction pipeline of the \trajdif model. Here we represent the data as a two-channel one-dimensional signal; the two channels in the input and output correspond to the two dimensions of the position coordinate. Right: Illustration of the conditional denoising process inside a diffusion block. The input and output are conditioned on the observed trajectory and the predicted intents. The U-net-encoded bottleneck features are attended across the channels and decoded. The output is then conditioned by the observed trajectory history and the predicted waypoints.
     }
     \vspace{-0.2in}
     \label{fig:network_structure}
\end{figure*}

We define the trajectory prediction problem as follows: Given the observed trajectory of an agent, denoted as $X = (x_1, \dots, x_{T_o})$, where each $x_t \in \mathbb{R}^2$ represents the agent's 2D coordinates in timestep $t$ within the observed frames $T_o$, as well as the semantic map $M$ of the surrounding environment, our objective is to predict the future trajectory $Y = (y_{T_o+1}, \dots, y_{T_o+T_p})$ for subsequent frames $T_p$. Here, $y_t \in \mathbb{R}^2$ denotes the 2D coordinates of the agent within the same coordinate system as $X$. 

To achieve our objective, we formulate the trajectory prediction task as follows.
\begin{equation}
P(Y, G|X,M) = P(Y|G,X,M)P(G|X,M).
\label{model_foundation}
\end{equation}
Here, the trajectory distribution is conditioned on an inferred agent intent $G$; we assume this intent to be the predicted long-term goal and short-term way-points the agent should follow, which makes $G \subset Y$. We used an off-the-shelf goal predictor to infer $G$. To model the conditional trajectory distribution $P(Y|G,X,M)$, we introduce a novel conditional diffusion-based trajectory prediction method called \trajdif. This model takes advantage of the inferred intent of the agent, the observed trajectory, and a semantic scene map. It generates a trajectory that corresponds to the agent's intent, guided by a map-based guidance module that ensures that the generated trajectory adheres to environmental constraints.

In Section \ref{diffusion fundamentals}, we introduce the basic formulation of the diffusion model. Section \ref{condition_diffusion_traj_pred} introduces the formulation of the model \trajdif in terms of input/output representation, model formulation, model architecture, and training. Section \ref{map_guide} introduces a map-based guidance term which ensures that the sampled predictions adhere to environmental constraints. Finally, Section \ref{sampling} introduces the sampling process during test time to generate trajectory predictions using the \trajdif model.

\subsection{Fundamentals of Denoising Diffusion Model} \label{diffusion fundamentals}
The Diffusion Model \cite{ho2020denoising,sohl2015deep} models the data generation procedure as an iterative denoising process $P_{\theta}(\tau^{i-1}|\tau^i)$ for $i=N, \dots, 0$ starting from $\tau^N \sim \mathcal{N}(0, I)$ sampled from a standard Gaussian distribution and $\tau^0$ being the ground truth data instance. This process is the inverse of a forward procedure that gradually adds noise to the ground truth data instance based on a sequence of variance schedule hyperparameters $\alpha = \{\alpha_1,...,\alpha_N\}$ which can be written as
\begin{equation}
    q(\tau^t|\tau^{t-1}) \sim \mathcal{N}(\tau^t; \sqrt{\alpha_t}\tau^{t-1}, (1-\alpha_t)\mathbf{I}).
    \label{forward_noising}
\end{equation}

The reverse denoising model $P_{\theta}(\tau^{i-1}|\tau^i)$ is often modeled as

\begin{equation}
    P_{\theta}(\tau^{i-1}|\tau^i) \sim \mathcal{N}(\mu_{\theta}(\tau^i, i), \sigma^2_q(i)\mathbf{I}).
    \label{reverse_model}
\end{equation}

Here, the mean function $\mu_{\theta}(\tau^i, i)$ is often parameterized with a neural network structure that takes the output of the previous denoising step and an embedding for the current denoising step index, and $\sigma^2_q(i)$ is a constant function of the scheduling hyperparameters $\alpha$. We follow the definition in \cite{luo2022understanding} for this diffusion process.

\subsection{Conditional Diffusion Model for Trajectory Prediction}\label{condition_diffusion_traj_pred}

\paragraph{\textbf{Trajectory Representation.}}
We represent the input and output trajectories of the model $ \tau \in \mathbb{R}^{(T_o+T_p) \times 2}$ as the concatenation of the observed trajectory $X$ and the predicted trajectory $\hat{Y}$. Then, the trajectory output from denoising step $i$ will have the form
\begin{equation}
    \tau^i = (x_1, \dots, x_{T_o}, \hat{y}^i_{T_o+1}, \dots, \hat{y}^i_{T_o+T_p-1}, \hat{y}_{T_o+T_p})'.
\end{equation}

There are two notions of timesteps, the index of denoising steps and the number of trajectory frames. We will use superscript to represent the former and subscript to represent the latter. This means that $\hat{y}^i_{T_o+1}$ represents the agent's 2D coordinates in the $T_o+1$ frame from the output of the $i$th denoising step.

We assume the predicted intent $G$ consists of a predicted goal $\hat{y}_{T_o+T_p}$ and $S$ waypoints $\hat{y}_{w_1} \dots \hat{y}_{w_S}$, where $w_s \in \{T_o+1, \dots ,T_o+T_p-1\}$ for all $s \in \{1,\dots ,S\}$, of the form
\begin{equation}
    G = \{\hat{y}_{w_1} \dots \hat{y}_{w_S}, \hat{y}_{T_o+T_p}\}.
\end{equation}

\paragraph{\textbf{Model Formulation.}}

Inspired by the image-inpainting task presented in \cite{sohl2015deep}, our \trajdif model $P(Y|G,X,M)$ considers the trajectory prediction task as an interpolation for the observed trajectory history and the predicted waypoints and the end goal. We use the diffusion model to perform iterative denoising on the coordinates between the observed history and the predicted goal and waypoints. Our model has the form
\begin{equation}\small
    P(Y|G,X,M) = P(\tau^0) =  \int P(\tau^N)\prod_{i=0}^{N-1}P(\tau^i|\tau^{i+1})d\tau^{1:N}.
\end{equation}

For each denoising step $i$, we fix the elements in $\tau^i$ corresponding to the observed trajectory with the observed coordinate sequence $(x_1, \dots, x_{T_o})$ and the elements corresponding to the predicted goal and the waypoints in the predicted intent $G$ to obtain a noisy conditioned trajectory $\tau^{i'}$. We then feed the $\tau^{i'}$ into the next denoising step $i-1$ and sample the next denoised trajectory $\tau^{i-1}$ based on the distribution in \eqref{reverse_model}. \autoref{fig:network_structure} illustrates how the input to each diffusion step is conditioned by the observed trajectory and the predicted agent intent.

\paragraph{\textbf{Architecture.}}
Following \cite{luo2022understanding}, we further parameterize the mean function $\mu_{\theta}(\tau^i, i)$ as 
\begin{equation}
    \mu_{\theta}(\tau^i, i) = \mu(\tau_{\theta}(\tau^i, i), \alpha),
    \label{repamarmeterization}
\end{equation}
where $\tau_{\theta}(\tau^i)$ is a neural network that predicts the ground truth trajectory with the noisy trajectory as its input. We use a U-Net-based \cite{ronneberger2015u} structure along with an attention module \cite{vaswani2017attention} for $\tau_{\theta}(\tau^i)$. We considered the two-dimensional trajectory in the form of a two-dimensional coordinate sequence as a 2-channel 1-dimensional image with only one temporal dimension. Therefore, we use groups of 1-dimensional convolutional neural network blocks along with residual connections for the encoding and decoding modules for the U-Net structure, following the model architecture design in \cite{janner2022diffuser}. The encoded U-Net feature, which has a dimension of $C \times W$, does not contain any information across the channels during the encoding process. Therefore, we feed the encoded feature into a cross-channel attention layer to resolve this issue, as suggested in \cite{janner2022diffuser}. The cross-attended feature is the input into the decoding blocks to generate the denoised trajectory. \autoref{fig:network_structure} illustrates the structure introduced here.

\paragraph{\textbf{Training.}}
Like standard denoising diffusion models, to train \trajdif, we minimize the KL divergence between our denoising model $P_{\theta}(\tau^{i-1}|\tau^i)$ and the ground-truth denoising distribution $q(\tau^{i-1}|\tau^t, \tau^0)$ for each denoising step $i \in [1,N]$. This KL divergence becomes the $L_2$ norm between the predicted mean function $ \mu_{\theta}(\tau^i, i)$ and the ground-truth mean function $\Tilde{\mu}(\tau^t, \tau^0)$. Based on the reprameterization in (\ref{repamarmeterization}), we can further simplify this and form our loss function as 
\begin{equation}\small
    \mathcal{L} = \mathbb{E}_{i \sim U\{1,N\}}\left[\mathbb{E}_{q(\tau^i|\tau^{i-1})}\left[\frac{\lambda(\alpha)}{2 \sigma^2_q(t)}\|\tau^{\theta}(\tau^i, t)-\tau^0\|^2_2\right]\right].
\end{equation}
The $\lambda(\alpha)$ is a function of the predefined noise schedule hyperparameter $\alpha$. The complete derivation follows the review \cite{luo2022understanding} and the original DDPM paper \cite{ho2020denoising}. We also provide a brief derivation in the Supplementary Material.

\subsection{Map-based Gradient Guidance Module}\label{map_guide}
To achieve the goal of generating trajectory predictions that strictly adhere to environmental constraints, we introduce a novel map-based gradient guidance term for the denoising trajectory prediction process, illustrated as the Map Based Guidance module in \autoref{fig:network_structure}. Given the semantic map $M$, we first extract the binary navigability map $M_b$ and perform the distance transform $D(M_b)$ to obtain the distance map $M_d$.  Each element $M_d(i,j)$ represents the distance to the closest navigable pixel, as demonstrated on right of \autoref{fig:fig1}. We then calculate the image gradient $\nabla M_d$ with respect to each pixel coordinate.

Directly using the gradient to guide the denoised trajectory may cause the coordinates of different frames to drift toward opposing directions, causing impossible predictions. Therefore, we propose an algorithm to guide the trajectory iteratively through the trajectory sequence. For each frame in the trajectory, we will perform a gradient descent using the image gradient $\nabla M_d$ to ensure that the coordinates of the current frame end up in a navigable area; we then also update all the positions of the later frames relative to the updated location of the current frame, as illustrated in \autoref{alg:map_guide}. We continue to perform this update iteratively for all time steps $\{t |t=T_o+1, \dots, T_o+T_p \}$.

\begin{algorithm}[t]
\vspace{0.5em}
\caption{Map-based Guidance $\mathcal{H}(\nabla M_d, \tau^i)$}\label{alg:map_guide}
\KwIn{denoised trajectory $\tau^i$}
\KwOut{map-based guidance term}
$\tau^{i*} \leftarrow \tau^i$\;

\For{$f = T_o+1$ to $T_o+T_p$}{
    \For{$k = K, \ldots, 1$}{\tcp{K times of this gradient descent step}
        $\tau^{i*}[f:] \gets \tau^{i*}[f:] + \nabla M_d(\tau^{i*}[f])$\;
    }
}
\Return $\tau^{i*} - \tau^i$
\end{algorithm}

\subsection{Test Time Sampling}\label{sampling}
During test time, we assume that an agent has an observed history $X$, the semantic scene map $M$, and $K$ sets of predicted agent intent $G=\{G_1,\dots, G_K\}$. For each set $G_k$, we sample a trajectory prediction through the conditional denoising diffusion process guided by the map-based guidance $\mathcal{H}$ defined in Section \ref{map_guide}. The pseudocode for the guided prediction pipeline is given in the supplementary.

\section{Experiments}\label{sec:experiment}

Section \ref{preliminaries} introduces the datasets, evaluation metrics, and benchmark settings. Section \ref{quantitative}  quantitatively compares the SOTA models and the \trajdif model. Section \ref{qualitative} performs qualitative analysis for the SOTA models and our \trajdif model. In section \ref{abalation}, we present an ablation study for the map-based guidance module. We also performed an analysis of the inference speed compared to other SOTA models, included in the supplementary section G. 

\subsection{Preliminaries.} \label{preliminaries}
\paragraph{\textbf{Datasets.}} We use two publicly available datasets for our benchmark experiment. The \textbf{nuScenes} dataset, first introduced in \cite{caesar2020nuscenes}, is a large-scale vehicle trajectory dataset, consisting of 1000 driving scenes and provides HD semantic maps. There are multiple benchmark configurations for this dataset and we follow \cite{yuan2021agentformer} and \cite{lee2022muse} and used their training and testing splits for the nuScenes prediction challenge. The \textbf{PFSD} dataset introduced in \cite{lee2022muse} features simulated trajectories within a group of synthetic path-finding environment layouts proposed in \cite{sohn2020laying}. The non-navigable areas are designed to be more complex for navigation. Despite having different agent types (vehicle and pedestrian, respectively), both benchmark offers strict navigability constraint, making them great benchmark platforms testing the models' ability generating realistic predictions.

\paragraph{\textbf{Metrics.}} For the benchmark evaluation, we compute the minimum average displacement error $\mathbf{ADE_k}$ and the final displacement error $\mathbf{FDE_k}$ of the $K$ trajectory samples for each agent compared with the ground truth trajectory. We also adopt the Kernel Density Estimated-based Negative Log Likelihood (\textbf{KDE NLL}) proposed in \cite{ivanovic2019trajectron}, which evaluates the fit of the model. We evaluate the environment understanding of the prediction models using the Environmental Collision-Free Likelihood (\textbf{ECFL}) proposed in \cite{sohn2021a2x}, which measures the probability that a predicted trajectory free of environmental collision. Finally, we quantify the diversity of the prediction outputs using the Multiverse Entropy \textbf{(MVE)} proposed in \cite{sohn2021a2x} quantifying the diversity of the trajectory predictions. We provides a detailed definition of each metric in the supplementary sections of the paper.

\paragraph{\textbf{Implementation Details.}} 
We use the CVAE-based Macro-stage of the MUSE-VAE model \cite{lee2022muse} to predict intent of an agent. We also point out that, for the following experiments, \trajdif and MUSE-VAE share the same sets of predicted waypoint sets. Other implementation details are presented in the supplementary.

\subsection{Quantitative Analysis} \label{quantitative}
For the quantitative experiment, we perform the trajectory prediction task on the two datasets mentioned above. We compared the performance of \trajdif against the Trajectron++ (T++) \cite{salzmann2020trajectron++}, AgentFormer (AF) \cite{yuan2021agentformer}, Y-net \cite{mangalam2021goals}, MUSE-VAE (MUSE) \cite{lee2022muse} and motion indeterminacy diffusion (MID) \cite{Gu_2022_CVPR}. Both PFSD and nuScenes provide rasterized global scene maps, and we provide a local view of the maps for all the methods benchmarked following the experiment in \cite{lee2022muse}. The original MID model does not utilize map information, so we also compare with a modified MID that incorporates a map embedding for fair comparison. Since MID uses T++'s encoder for observed trajectories and social interactions, we use T++'s map encoder for the modified MID. We point out this is different from our map guidance module, but an adaption based on the design choice of MID model itself. We trained all models from scratch except for MUSE and Agentformer, where we used their provided pre-trained weight on the nuScenes dataset and we used MUSE-VAE's pretrained weight on PFSD dataset. Some benchmark models consider multi-agent settings, using context from other agents to condition independent or joint predictions of all agents' trajectories in the scene. We argue that such a difference does not significantly affect our conclusions, and we provide a more detailed discussion in the supplementary material.

\begin{table}[htb]
\vspace{-1.5\baselineskip}
    \centering
    \caption{Quantative Results on PFSD and nuScenes datasets}
    \begin{subtable}{.45\linewidth}
    \caption{Results on PFSD with $K=20$ with $t_{obs}=3.2$s (8 frames) and $t_{pred}=4.8$s (12 frames). Errors are in meters. The best performance is boldfaced and the 2nd place is marked as blue. Numbers in parenthesis indicate the ranking for the score. 
    }
    \resizebox{\columnwidth}{!}{
    \begin{tabular}{l|ccccc}
        \toprule
         Model & ADE $\downarrow$ & FDE $\downarrow$ & NLL $\downarrow$ & ECFL $\uparrow$ & MVE $\uparrow$ \\
         \midrule
         T++ &  0.20 (7)        & 0.42 (7) & 2.24 (7) & 85.00 (7) & \textbf{1.13} (1) \\
         AF & 0.11  (6)         & 0.17 (5) & \textbf{1.93} (1)& 93.76 (4) & 0.67 (7) \\
         Y-net & 0.07 (3)       & 0.12 (3) & 1.98 (3) & 94.16 (3) & 0.79 (6) \\
         MUSE & \textbf{0.05} (1)& \textbf{0.09} (1) & \textcolor{blue}{1.95} (2) & \textcolor{blue}{97.08} (2) & 0.92 (4) \\
         MID & 0.09 (4) & 0.16 (4) & 2.00 (5) & 88.72 (6) & 0.93 (3) \\
         MID w/Map & 0.10 (5) & 0.19 (6) & 2.00 (5) & 90.41 (5) & 0.86 (5)  \\
         \rowcolor{red!20} \textbf{\trajdif} & \textcolor{blue}{0.06} (2)& \textbf{0.09} (1) & 1.98  (3) & \textbf{99.62} (1) & \textcolor{blue}{1.08} (2) \\
         \bottomrule
    \end{tabular}
    }
    \label{tab:PFSD_result}
    \end{subtable}\hspace{0.06\linewidth}%
    \begin{subtable}{.45\linewidth}
    \small
    \centering
    \caption{Result on nuScenes with $K=5$ and $K=10$ with $t_{obs}=2$s (4 frames) and $t_{pred}=6$s (12 frames). Errors are in meters. The best performance is boldfaced and the 2nd place is marked as blue. Numbers in parenthesis indicate the ranking for the score. 
    }
    \resizebox{1.05\columnwidth}{!}{
    \begin{tabular}{c|lccccc}
        \toprule
         K &  Model & ADE $\downarrow$ & FDE $\downarrow$ & NLL $\downarrow$ & ECFL $\uparrow$ & MVE $\uparrow$ \\ \midrule
         \multirow{7}{*}{5} &T++ &  2.51 (7) & 5.57 (6) & 11.66 (7) & 81.66 (4) & 0.46 (6) \\
         &AF & 1.86 (4) & 3.89 (4) & 6.94 (3) & 84.66 (3) & 0.38 (7) \\
         &Y-net &\textcolor{blue}{1.63} (2) & 2.86 (3) & 7.13 (4) & 76.61 (5) & \textcolor{blue}{0.68} (3) \\
         &MUSE & \textbf{1.37} (1) & \textcolor{blue}{2.84} (2) & \textbf{5.76} (1) & \textcolor{blue}{89.30} (2) &  0.65 (4)  \\
         &MID & 2.38 (5) & 5.54 (5) & 9.33 (5) & 69.23 (6) & \textbf{0.81} (1) \\
         &MID w/Map & 2.42 (6) & 5.61 (6) & 9.51 (6) & 68.72 (7) & \textbf{0.81} (1) \\
         &\cellcolor{red!20}\textbf{\trajdif} & \cellcolor{red!20}1.67 (3) & \cellcolor{red!20}\textbf{2.73} (1) & \cellcolor{red!20}\textcolor{blue}{6.85} (2) &\cellcolor{red!20} \textbf{99.15} (1) & \cellcolor{red!20}0.61 (5) \\ \midrule
         \multirow{7}{*}{10} &T++ &  1.92 (5) & 4.01 (5) & 8.20 (7) & 81.25 (4) & 0.57 (6)  \\
         &AF & 1.45 (4) & 2.86 (4) & 5.67 (4) & 84.26 (3) & 0.42 (7)  \\
         &Y-net & \textcolor{blue}{1.32} (1) & \textcolor{blue}{2.05} (2) & 5.60 (3) & 70.71 (5) & \textbf{1.03} (3) \\
         &MUSE & \textbf{1.10} (1) & 2.11 (3) & \textbf{4.61} (1) & \textcolor{blue}{89.26} (2) & 0.79 (4) \\
         &MID & 1.93 (6) & 4.29 (7) & 7.42 (6) & 68.97 (6) & \textcolor{blue}{1.00} (2) \\
         &MID w/Map & 1.96 (7) & 4.28 (6) & 7.41 (5) & 68.40 (7) & \textcolor{blue}{1.00} (2) \\
         &\cellcolor{red!20}\textbf{\trajdif} & \cellcolor{red!20}1.41 (3) & \cellcolor{red!20}\textbf{2.02} (1) & \cellcolor{red!20}\textcolor{blue}{5.33} (2) & \cellcolor{red!20}\textbf{99.08} (1) & \cellcolor{red!20}0.74 (5) \\ \bottomrule
    \end{tabular}
    }
    \label{tab:nuScenes_result}
    \end{subtable}
    \vspace{-2\baselineskip}
\end{table}

\paragraph{\textbf{PFSD dataset.}} \autoref{tab:PFSD_result} summarizes the results in the PFSD dataset. For the PFSD dataset, we use 3.2 second (8 frames) observations and predict 4.8 second (12 frames) into the future. We chose to sample $K = 20$ samples for the PFSD dataset to consider the inherent multimodal nature of the human trajectory. For the $ADE_{20}$ score, \trajdif was able to achieve the second best among all the benchmarked models. For the $FDE_{20}$ score, \trajdif achieves the overall best along with the MUSE-VAE model. The two displacement errors measure the prediction accuracy of the best sample and we have shown that \trajdif can achieve SOTA performance in terms of these two metrics. For the KDE NLL metric, \trajdif also matches the SOTA methods, meaning that the K samples generated could reflect the distribution of the ground truth data distribution.  \trajdif model is capable of surpassing all SOTA models in the ECFL experiment. The \trajdif model can achieve an almost perfect prediction in terms of complying with the environmental constraint. This indicates that given sufficient predicted waypoints, \trajdif is capable of generating predictions that are accurate and realistic. For the MVE, \trajdif is ranked second. This indicates that \trajdif is capable of generating diverse trajectory predictions. Compared to MID, another diffusion-based model, \trajdif achieved better performance in all metrics. For the PFSD data set, after adding map embedding, MID achieves a lower MVE score, causing the predictions to be less diverse. This is also demonstrated in the qualitative analysis.

\paragraph{\textbf{nuScenes dataset.}} For the nuScenes dataset, we follow the configurations of previous works, observing 2 seconds (4 frames) of past trajectories and predicting 6 seconds (12 frames) into the future. We experiment with two $K$ settings for this dataset. \autoref{tab:nuScenes_result} shows the result. For the $K=5$ case, our model was able to achieve the best $FDE_5$ score and matches the SOTA $ADE_5$ performance. This indicates that the model can take advantage of the accurate long-term prediction and generate a reasonable and accurate trajectory. For the KDE NLL, \trajdif is able to achieve the second best. For the ECFL score, \trajdif can achieve the best among all models and achieve an increase of almost 10\%. For case $K=10$, all models achieve better displacement error performance, with MUSE having the best $ADE_{10}$ score and \trajdif having the best $FDE_{10}$ score. \trajdif was also able to achieve SOTA performance on the KDE NLL metric. For the ECFL, \trajdif again achieves the best among the benchmark models and achieves almost perfect predictions it terms of following the environmental constraint. For the MVE metric, \trajdif is ranked 5th overall. However, compared to the top three ranking MID, Y-net and MUSE-VAE, \trajdif is capable of generating more feasible predictions while still having comparable MVE values. On the other hand, the top-ranking Y-net and MID sacrifice the feasibility in terms of ECFL to achieve more diversity, causing most of their prediction to be unrealistic. Although MUSE-VAE ranked the highest on average, its generated trajectories often violated environmental constraints, making unrealistic predictions, as can be seen in the qualitative analysis in the next section. The MID model struggled to perform on the vehicle-focused nuScenes dataset. Despite ranking first for the MVE metric for diversity, the model fails to produce accurate and feasible trajectory predictions. The experiments show that diffusing from raw Gaussian noise and using only trajectory history embedding as conditions is not as effective compared with our setting of interpolation via diffusion. 

\begin{figure*}
    \centering
    \begin{subfigure}{1\linewidth}
    \includegraphics[width=1\linewidth]{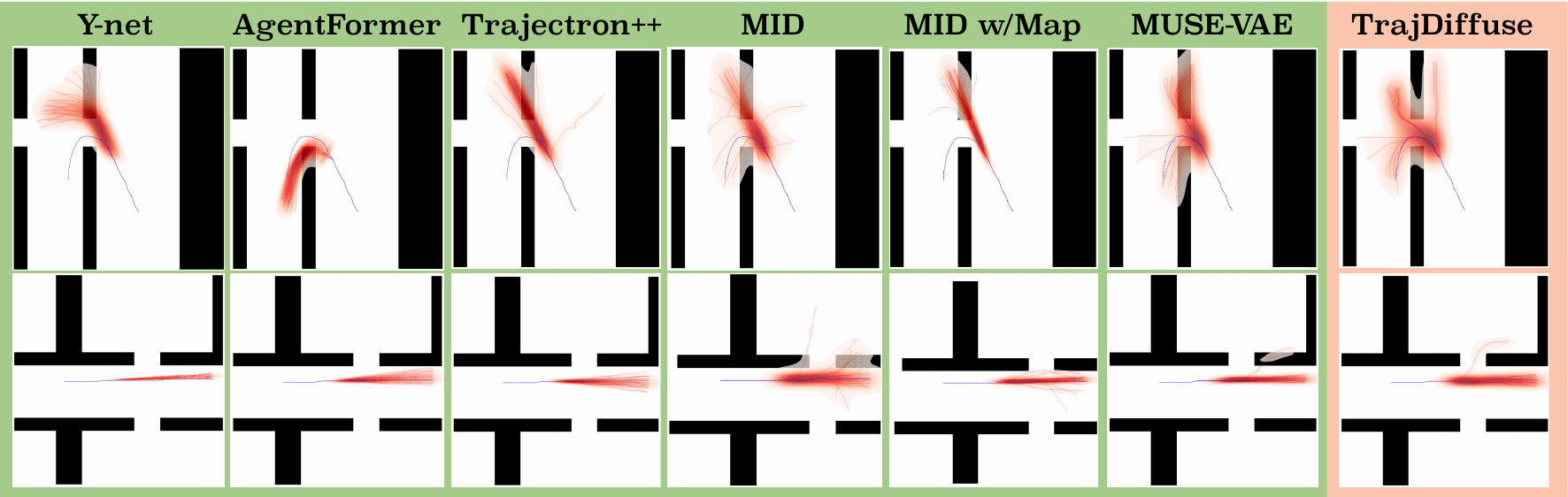}
    \caption{Visualizations for PFSD with $K= 20$}
    \label{fig:pfsd_visual}
    \end{subfigure}

    \begin{subfigure}{1\linewidth}
    \includegraphics[width=1\linewidth]{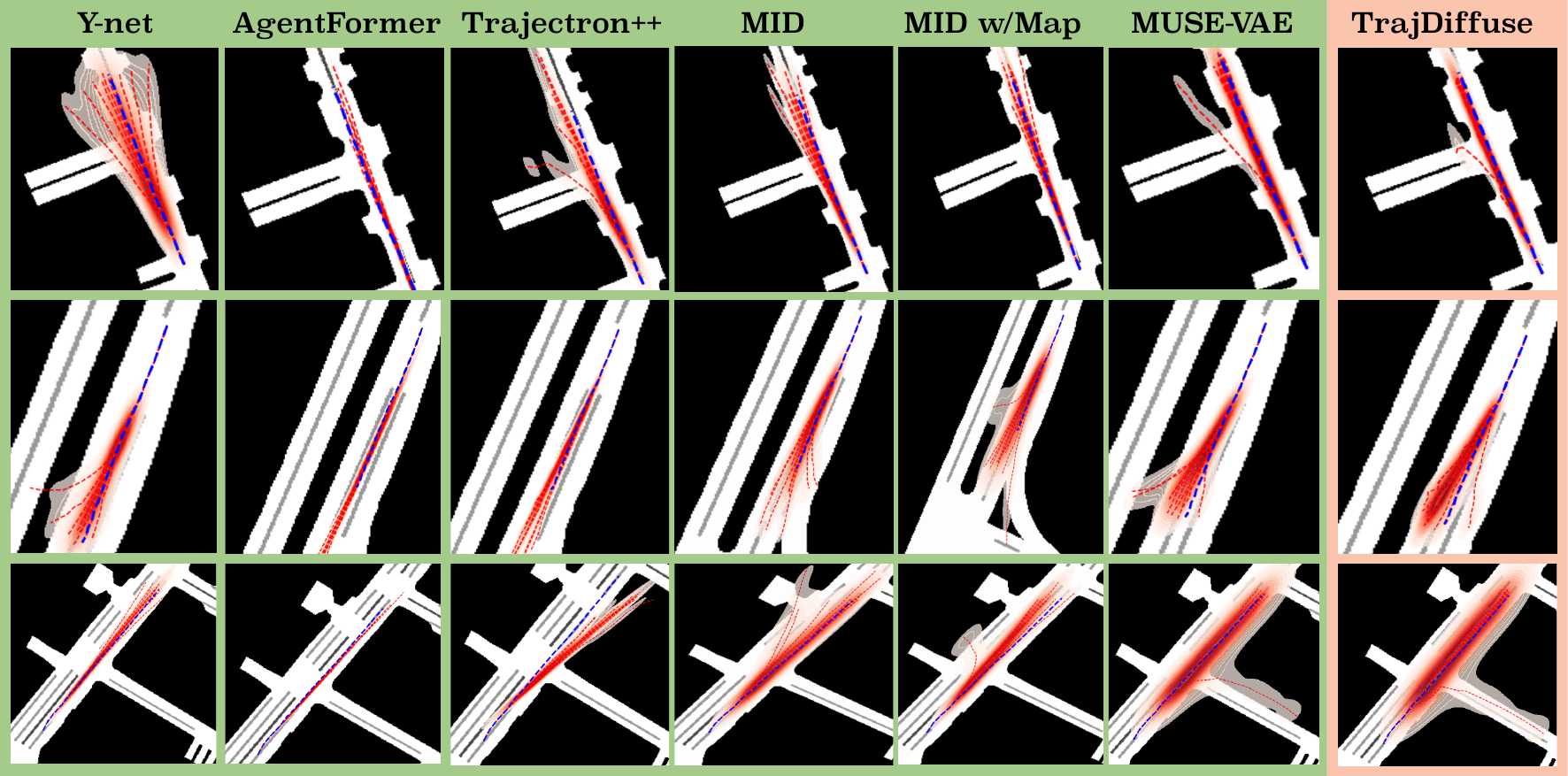}
    \caption{Visualizations for nuScenes with $K=10$}
    \label{fig: nuscenes_visual}
    \end{subfigure}
     \caption{Visualizations for qualitative analysis on PFSD and nuScenes datasets. Each column contains visualizations of an agent's trajectories predicted by the model indicated at the top of the column. Each row corresponds to the agent with identical initial conditions and identical prior motion history. 
     {\color{blue} Blue} dashed lines denote the ground-truth trajectories. {\color{red} Red} dashed lines are predicted trajectories.}
\end{figure*}

\subsection{Qualitative Analysis}\label{qualitative}

We provide a qualitative analysis with visualizations of the prediction output of each model to better illustrate the behavior and characteristics of these models. 

\autoref{fig:pfsd_visual} shows two instances of the PFSD experiment. For the PFSD dataset, the \trajdif model is capable of generating diverse paths that are also able to adhere to the environmental constraint. Here, we point out again that we use the same intent prediction as the MUSE-VAE output. Since MUSE-VAE does not directly use the predicted goal and waypoints, it will still generate trajectories that cause a collision with the environment. Our guided denoising inpainting approach is able to address this issue while still maintaining the prediction accuracy. Y-net test time sampling trick helps the model generate a diverse set of predictions; however, this sampling process is not data-driven and generates trajectories that ignore the environmental constraint. The AgentFormer model is able to generate trajectories that cause no collision with the environment in one of the instances, however, the outputs lack diversity. For the other instance, AgentFormer is able to predict the correct intent of the agent; however, the trajectories all violate the environmetnal constaint and the predictions also lack diversity. Trajectron++ model suffers a lack of diversity issue in one of the cases shown here, and in the other instance the model also generates many trajectories which ignore the environmental constraint. The MID model (in both settings) behave similarly to the Trajectron++, as they share the same encoder structure. The model is able to achieve better environmental understanding with the addition of map embedding; however, it still produces trajectories violating the map constraints, and the addition of map embedding also causes the model to produce less diverse predictions in both instances.

\autoref{fig: nuscenes_visual} presents three instances from the nuScenes dataset. Due to the dataset's focus on vehicle trajectory in traffic scenarios, the environmental constraint is stricter, offering limited navigable areas. Here, the \trajdif model is capable of producing accurate and diverse trajectories and staying within navigable areas. \trajdif and MUSE-VAE also share the predicted waypoints in this experiment. However, we see here that MUSE will generate off-road trajectories, which are likely due to inaccurate intent predictions. Meanwhile, the \trajdif model with its map-based guidance is able to correct those trajectories and makes reasonable predictions. For Y-net outputs, the disadvantage of the test-time sampling trick is shown in this kind of narrow environment. Y-net outputs have shown a trade-off between diversity and environmental compliance, where many outputs go off-road. The sequence-to-sequence setting of AgentFormer and Trajectron++ was able to generate more output that stays on-road. However, these predictions often fail to reach the ground truth goal and overshoot beyond the truth trajectory, causing the prediction accuracy problem. MID also suffers from this issue of inaccurate predictions, and the model also often goes off road given the narrower navigable areas. With the help of map embedding, the model does have lower environmental violation; however, it still suffers in accuracy, and the model also produces less diverse outputs for these three instances.

From the qualitative analysis, we have shown that the \trajdif offers a better solution for using the predicted intent than the MUSE-VAE model and renders more realistic and diverse predictions than MID Y-Net, AgentFormer and Trajectron++ models.

\subsection{Ablation Study}
\label{abalation}

\autoref{tab:abalation} presents an ablation study of the map-based guidance sampling on the nuScenes dataset with $K=10$ samples. We note that the map-based guidance not only improves the ECFL of the model output; it also improves the accuracy metrics. The MVE value decreases slightly; however the MVE metric should be considered along with other metrics and this indicates the guidance term helps the model to generate higher quality trajectory predictions while maintaining the diversity. We also note that using the heatmap based MUSE Macro-stage for intent prediction indeed helps the \trajdif achieves a good starting base line for ECFL, however, we see that using the same sets of predicted waypoints, \trajdif without map guidance still achieves better ECFL score than MUSE and both \trajdif settings also achieve better FDE score compare with MUSE. This is due to the non-autoregressive setting of our diffusion based interpolation, which directly uses the environmentally-compliant waypoints as the final output. On the other hand, MUSE-VAE's micro-stage only uses these predicted waypoints as reference for its auto-regressive RNN based prediction module and generate brand new trajectory predictions that often fails to maximizes the advantage of the environment-aware Macro-stage predictions.

We provide another ablation study in \autoref{tab:abalation2} on the number waypoints used for prediction on the nuScenes dataset. 
With only one waypoint (final goal), the model yields slightly decreased performance. However, for other choices, the model performance increases and does not drastically vary. Hence, the model needs both the intermediate waypoints and the goal point, but it is robust against a different number of waypoints. We chose 3 waypoints following MUSE-VAE.

\begin{table}[htb]
    \centering
    \vspace{-1.5\baselineskip}
    \caption{Two abalation studies}
    \vspace{-1.0\baselineskip}
    \begin{subtable}{.45\linewidth}
    \small
    \centering
    \caption{Ablation study of the map-based guidance on nuScenes dataset with $K = 10$.}
    \resizebox{\columnwidth}{!}{
    \begin{tabular}{l|ccccc}
         \toprule
         Model & ADE $\downarrow$ & FDE $\downarrow$ & KDE NLL $\downarrow$ & ECFL $\uparrow$ & MVE $\uparrow$ \\ \hline
         MUSE-VAE & 1.10 & 2.11 & 4.61 & 89.26 & 0.79 \\
        \rowcolor{red!5}\trajdif w/o Guidance & 1.42 & 2.04 & 5.40 & 92.07 & 0.77 \\
         \rowcolor{red!20} \trajdif w Guidance & 1.42 & 2.02 & 5.33 & 99.08 & 0.75 \\
         \bottomrule
    \end{tabular}
    }
    \label{tab:abalation}
    \end{subtable}\hspace{0.06\linewidth}%
    \begin{subtable}{.45\linewidth}
    \small
    \centering
        \caption{Ablation study for different number of waypoints on nuScenes with $K=5$ with $t_{obs}=2$s (4 frames) and $t_{pred}=6$s (12 frames). Errors are in meters; map-guide vs. w/o map-guide}
    \vspace{-0.1in}
    \resizebox{\columnwidth}{!}{
    \begin{tabular}{c|ccccc}
         \#Waypoints & ADE $\downarrow$ & FDE $\downarrow$ & NLL $\downarrow$ & ECFL $\uparrow$ & MVE $\uparrow$\\ \hline 
         1 & 1.74/1.77 & 2.75/2.78 & 7.18/7.30 & 99.16/89.12 & 0.57/0.60 \\
         2 & 1.68/1.69 & 2.76/2.80 & 7.12/7.22 & 99.15/91.64 & 0.60/0.62 \\
         3 & 1.67/1.68 & 2.73/2.76 & 6.85/6.94 & 99.15/92.10 & 0.61/0.64 \\
         4 & 1.66/1.67 & 2.77/2.79 & 7.15/7.20 & 99.02/92.24 & 0.60/0.62 \\
    \bottomrule
    \end{tabular}
    }
    \label{tab:abalation2}
    \end{subtable}
    \vspace{-2\baselineskip}
\end{table}

\section{Discussion and Conclusion} \label{sec:conclusion}

In this paper, we introduced \trajdif, a guided conditional diffusion model for trajectory prediction. By framing the prediction task as denoising interpolation of the observed trajectory and predicted waypoints, our model is able to achieve a SOTA performance in accuracy and diversity measure while surpassing the existing method's ability to follow the critical environmental constraints. 

Several other future directions can be considered. Our \trajdif framework makes it possible to integrate other dynamic scene elements and HD-maps that have become more prevalent in recent years. Alternate approaches that incorporate the intent of other agents in the scene to model agent-agent interactions may similarly and readily benefit from the \trajdif framework.

Our method leverages the predicted agent intent in the form of trajectory waypoints; in this work, we employ a CVAE-based backbone of the MUSE-VAE model \cite{lee2022muse} for that task. Although we demonstrated that the feasibility of the \trajdif prediction exceeds that of MUSE-VAE, other waypoint prediction models could be used to create successful pipeline designs, opening an interesting direction for further investigation.

\paragraph{\textbf{Acknowledgement}}This project is supported in part by National Science Foundation Grant Awards \#1955404 and \#1955365.
\small
\bibliographystyle{splncs04}
\bibliography{main}

\begin{thebibliography}{10}
\providecommand{\url}[1]{\texttt{#1}}
\providecommand{\urlprefix}{URL }
\providecommand{\doi}[1]{https://doi.org/#1}

\bibitem{alahi2016social}
Alahi, A., Goel, K., Ramanathan, V., Robicquet, A., Fei-Fei, L., Savarese, S.: Social lstm: Human trajectory prediction in crowded spaces. In: Proceedings of the IEEE conference on computer vision and pattern recognition. pp. 961--971 (2016)

\bibitem{caesar2020nuscenes}
Caesar, H., Bankiti, V., Lang, A.H., Vora, S., Liong, V.E., Xu, Q., Krishnan, A., Pan, Y., Baldan, G., Beijbom, O.: nuscenes: A multimodal dataset for autonomous driving. In: Proceedings of the IEEE/CVF conference on computer vision and pattern recognition. pp. 11621--11631 (2020)

\bibitem{chai2019multipath}
Chai, Y., Sapp, B., Bansal, M., Anguelov, D.: Multipath: Multiple probabilistic anchor trajectory hypotheses for behavior prediction. arXiv preprint arXiv:1910.05449  (2019)

\bibitem{10.1109/ICRA46639.2022.9812253}
Gilles, T., Sabatini, S., Tsishkou, D., Stanciulescu, B., Moutarde, F.: Gohome: Graph-oriented heatmap output for future motion estimation. In: 2022 International Conference on Robotics and Automation (ICRA). p. 9107–9114. IEEE Press (2022)

\bibitem{goodfellow2014generative}
Goodfellow, I., Pouget-Abadie, J., Mirza, M., Xu, B., Warde-Farley, D., Ozair, S., Courville, A., Bengio, Y.: Generative adversarial nets. Advances in neural information processing systems  \textbf{27} (2014)

\bibitem{Gu_2022_CVPR}
Gu, T., et~al.: Stochastic trajectory prediction via motion indeterminacy diffusion. In: CVPR (June 2022)

\bibitem{gupta2018social}
Gupta, A., Johnson, J., Fei-Fei, L., Savarese, S., Alahi, A.: Social gan: Socially acceptable trajectories with generative adversarial networks. In: Proceedings of the IEEE conference on computer vision and pattern recognition. pp. 2255--2264 (2018)

\bibitem{ho2020denoising}
Ho, J., Jain, A., Abbeel, P.: Denoising diffusion probabilistic models. Advances in neural information processing systems  \textbf{33},  6840--6851 (2020)

\bibitem{ivanovic2019trajectron}
Ivanovic, B., Pavone, M.: The trajectron: Probabilistic multi-agent trajectory modeling with dynamic spatiotemporal graphs. In: Proceedings of the IEEE/CVF International Conference on Computer Vision. pp. 2375--2384 (2019)

\bibitem{janner2022diffuser}
Janner, M., Du, Y., Tenenbaum, J., Levine, S.: Planning with diffusion for flexible behavior synthesis. In: International Conference on Machine Learning (2022)

\bibitem{Jiang_2023_CVPR}
Jiang, C.{\textquotedblleft}., et~al.: Motiondiffuser: Controllable multi-agent motion prediction using diffusion. In: CVPR (2023)

\bibitem{kosaraju2019social}
Kosaraju, V., Sadeghian, A., Mart{\'\i}n-Mart{\'\i}n, R., Reid, I., Rezatofighi, H., Savarese, S.: Social-bigat: Multimodal trajectory forecasting using bicycle-gan and graph attention networks. Advances in Neural Information Processing Systems  \textbf{32} (2019)

\bibitem{lee2022muse}
Lee, M., Sohn, S.S., Moon, S., Yoon, S., Kapadia, M., Pavlovic, V.: Muse-vae: multi-scale vae for environment-aware long term trajectory prediction. In: Proceedings of the IEEE/CVF Conference on Computer Vision and Pattern Recognition. pp. 2221--2230 (2022)

\bibitem{li2023multi}
Li, Z., et~al.: A multi-modal vehicle trajectory prediction framework via conditional diffusion model: A coarse-to-fine approach. Knowledge-Based Systems  \textbf{280},  110990 (2023)

\bibitem{luo2022understanding}
Luo, C.: Understanding diffusion models: A unified perspective. arXiv preprint arXiv:2208.11970  (2022)

\bibitem{mangalam2021goals}
Mangalam, K., An, Y., Girase, H., Malik, J.: From goals, waypoints \& paths to long term human trajectory forecasting. In: Proceedings of the IEEE/CVF International Conference on Computer Vision. pp. 15233--15242 (2021)

\bibitem{mao2023leapfrog}
Mao, W., et~al.: Leapfrog diffusion model for stochastic trajectory prediction. In: CVPR (2023)

\bibitem{misra_2019_BMVC}
Misra, D.: Mish: A self regularized non-monotonic neural activation function. In: Proceedings of the British Machine Vision Conference (BMVC) (2019)

\bibitem{nichol2021glide}
Nichol, A., Dhariwal, P., Ramesh, A., Shyam, P., Mishkin, P., McGrew, B., Sutskever, I., Chen, M.: Glide: Towards photorealistic image generation and editing with text-guided diffusion models. arXiv preprint arXiv:2112.10741  (2021)

\bibitem{nichol2021improved}
Nichol, A.Q., Dhariwal, P.: Improved denoising diffusion probabilistic models. In: International Conference on Machine Learning. pp. 8162--8171. PMLR (2021)

\bibitem{pang2021trajectory}
Pang, B., Zhao, T., Xie, X., Wu, Y.N.: Trajectory prediction with latent belief energy-based model. In: Proceedings of the IEEE/CVF Conference on Computer Vision and Pattern Recognition. pp. 11814--11824 (2021)

\bibitem{rasul2021autoregressive}
Rasul, K., Seward, C., Schuster, I., Vollgraf, R.: Autoregressive denoising diffusion models for multivariate probabilistic time series forecasting. In: International Conference on Machine Learning. pp. 8857--8868. PMLR (2021)

\bibitem{ronneberger2015u}
Ronneberger, O., Fischer, P., Brox, T.: U-net: Convolutional networks for biomedical image segmentation. In: Medical Image Computing and Computer-Assisted Intervention--MICCAI 2015: 18th International Conference, Munich, Germany, October 5-9, 2015, Proceedings, Part III 18. pp. 234--241. Springer (2015)

\bibitem{salzmann2020trajectron++}
Salzmann, T., Ivanovic, B., Chakravarty, P., Pavone, M.: Trajectron++: Dynamically-feasible trajectory forecasting with heterogeneous data. In: Computer Vision--ECCV 2020: 16th European Conference, Glasgow, UK, August 23--28, 2020, Proceedings, Part XVIII 16. pp. 683--700. Springer (2020)

\bibitem{sohl2015deep}
Sohl-Dickstein, J., et~al.: Deep unsupervised learning using nonequilibrium thermodynamics. In: ICML. pp. 2256--2265. PMLR (2015)

\bibitem{sohn2015learning}
Sohn, K., Lee, H., Yan, X.: Learning structured output representation using deep conditional generative models. Advances in neural information processing systems  \textbf{28} (2015)

\bibitem{sohn2021a2x}
Sohn, S.S., Lee, M., Moon, S., Qiao, G., Usman, M., Yoon, S., Pavlovic, V., Kapadia, M.: A2x: An agent and environment interaction benchmark for multimodal human trajectory prediction. In: Proceedings of the 14th ACM SIGGRAPH Conference on Motion, Interaction and Games. pp.~1--9 (2021)

\bibitem{sohn2020laying}
Sohn, S.S., Zhou, H., Moon, S., Yoon, S., Pavlovic, V., Kapadia, M.: Laying the foundations of deep long-term crowd flow prediction. In: Computer Vision--ECCV 2020: 16th European Conference, Glasgow, UK, August 23--28, 2020, Proceedings, Part XXIX 16. pp. 711--728. Springer (2020)

\bibitem{song2020denoising}
Song, J., Meng, C., Ermon, S.: Denoising diffusion implicit models. arXiv preprint arXiv:2010.02502  (2020)

\bibitem{song2023consistency}
Song, Y., Dhariwal, P., Chen, M., Sutskever, I.: Consistency models. arXiv e-prints pp. arXiv--2303 (2023)

\bibitem{song2019generative}
Song, Y., Ermon, S.: Generative modeling by estimating gradients of the data distribution. Advances in neural information processing systems  \textbf{32} (2019)

\bibitem{vaswani2017attention}
Vaswani, A., Shazeer, N., Parmar, N., Uszkoreit, J., Jones, L., Gomez, A.N., Kaiser, {\L}., Polosukhin, I.: Attention is all you need. Advances in neural information processing systems  \textbf{30} (2017)

\bibitem{wang2022diffusion}
Wang, Z., Hunt, J.J., Zhou, M.: Diffusion policies as an expressive policy class for offline reinforcement learning. arXiv preprint arXiv:2208.06193  (2022)

\bibitem{Wu_2018_ECCV}
Wu, Y., He, K.: Group normalization. In: Proceedings of the European Conference on Computer Vision (ECCV) (September 2018)

\bibitem{yuan2021agentformer}
Yuan, Y., Weng, X., Ou, Y., Kitani, K.M.: Agentformer: Agent-aware transformers for socio-temporal multi-agent forecasting. In: Proceedings of the IEEE/CVF International Conference on Computer Vision. pp. 9813--9823 (2021)

\end{thebibliography}

\title{Supplementary for:\\TrajDiffuse: A Conditional Diffusion Model for Environment-Aware Trajectory Prediction}
\author{Qingze (Tony) Liu\inst{1} \and
Danrui Li\inst{1} \and
Samuel S. Sohn\inst{1} \and \\
Sejong Yoon\inst{2} \and
Mubbasir Kapadia \inst{1} \and
Vladimir Pavlovic\inst{1} 
}

\institute{Rutgers University, Piscataway, USA \\
\email{\{tony.liu; danrui.li; mubbasir.kapadia\}@rutgers.edu; \\
\{sss286; vladimir\}@cs.rutgers.edu} \\
\and The College of New Jersey, Ewing, USA \\
\email{yoons@tcnj.edu}}

\maketitle

\appendix


In \autoref{sec:section loss derivation}, we include a derivation of the loss function for \trajdif training. In \autoref{sec: implementation detail}, we report more details of implementation as a supplement to Section 4.1 of the main paper. In \autoref{metric sup}, we offer detailed definitions of the metrics used in the quantitative analysis in the main paper. In \autoref{sec:model choice}, we discuss our choice of SOTA models for the benchmark. In \autoref{sec: model setting}, we discuss the difference in the prediction and context settings of the benchmark models and provide a new benchmark result on the agent-agent collision rate for the PFSD and nuScenes datasets. In \autoref{sec:additional result}, we report qualitative results for additional instances of the two benchmark datasets. In \autoref{sec:inference speed}, we provide an analysis of inference speed of \trajdif compare against SOTA models. Finally, in \autoref{sec: limitation}, we highlight some limitations and failure cases of our \trajdif, as well as potential future directions.

\section{Training Loss Derivation}
\label{sec:section loss derivation}

As mentioned in Sec. 3 of the main paper, during diffusion model training, we gradually add noise to the ground-truth trajectory $\tau^0$ based on a sequence of variance schedule hyperparameters $\alpha = \{\alpha_1,...,\alpha_N\}$ which can be written as
\begin{equation}
    q(\tau^i|\tau^{i-1}) \sim \mathcal{N}(\tau^t; \sqrt{\alpha_i}\tau^{i-1}, (1-\alpha_i)\mathbf{I}),
    \label{forward_noising_supp}
\end{equation}
for $t = 1, \dots, N$. During training, we can derive the distribution for
\begin{equation}
    q(\tau^{i-1}|\tau^i, \tau^0) = \frac{q(\tau^i|\tau^{i-1}, \tau^0)q(\tau^{i-1}|\tau^0)}{q(\tau^i|\tau^0)}.
\end{equation}
The forward process follows the Markovian assumption by design; therefore, we know $q(\tau^i|\tau^{i-1}, \tau^0) = q(\tau^i|\tau^{i-1})$ from \autoref{forward_noising_supp}. For $q(\tau^i|\tau^0)$ and $q(\tau^{i-1}|\tau^0)$, it can be easily derived, where
\begin{equation}
    q(\tau^i|\tau^0) \sim \mathcal{N}(\sqrt{\Bar{\alpha}_i}, (1-\Bar{\alpha}_i)\mathbf{I}).
\end{equation}
This is true for $i = 1, \dots, n$ and we have $\Bar{\alpha}_i = \prod_{j=1}^i\alpha_j$. Refer to Equations 61 to 70 in \cite{luo2022understanding} for more detail. Therefore, we know that 
\begin{equation}
     q(\tau^{i-1}|\tau^i, \tau^0) \sim \mathcal{N}(\mu_q(\tau^i, \tau^0),  \sigma^2_q(i)\mathbf{I}),
\end{equation}
where $\sigma^2_q(i)$ is a constant function of noising schedule $\alpha$ (as derived in \cite{luo2022understanding} and \cite{ho2020denoising}). And we have
\begin{equation}
    \mu_q(\tau^i, \tau^0) = \frac{\sqrt{\alpha_i}(1-\Bar{\alpha}_i)\tau^i+\sqrt{\Bar{\alpha}_{i-1}}(1-\alpha_i)\tau^0}{1-\Bar{\alpha}_i}.
\end{equation}

For the reverse denoising model,
\begin{equation}
    P_{\theta}(\tau^{i-1}|\tau^i) \sim \mathcal{N}(\mu_{\theta}(\tau^i, i), \sigma^2_q(i)\mathbf{I}).
    \label{reverse_model}
\end{equation}
Following $\mu_q(\tau^i, \tau^0)$, a natural choice for the mean function $\mu_{\theta}(\tau^i, i)$ is then

\begin{equation}
    \mu_{\theta}(\tau^i, i) = \frac{\sqrt{\alpha_i}(1-\Bar{\alpha}_i)\tau^i+\sqrt{\Bar{\alpha}_{i-1}}(1-\alpha_i)\tau_{\theta}(\tau^i,i)}{1-\Bar{\alpha}_i}.
\end{equation}
Here, the neural network $\tau_{\theta}(\tau^i, i)$ estimates the ground-truth trajectory $\tau^0$ at each denoising step. 

During training, we minimize the KL divergence $D_{KL}$ between the true denoising distribution $q(\tau^{i-1}|\tau^i, \tau^0)$ and our modeled distribution $P_{\theta}(\tau^{i-1}|\tau^i)$ for each denoising step, $D_{KL} (q(\tau^{i-1}|\tau^i, \tau^0)|P_{\theta}(\tau^{i-1}|\tau^i))$. Since both distributions are Gaussian, the KL divergence between two Gaussian distributions has a closed form and we can simplify the KL divergence to be
\begin{equation}
    D_{KL} (q|P_{\theta}) = \frac{1}{2\sigma^2_q(i)}\frac{\Bar{\alpha}_{i-1}(1-\alpha_i)^2}{(1-\Bar{\alpha}_i)^2}\|\tau_{\theta}(\tau^i,i)-\tau^0\|_2^2
\end{equation}

Hence, our loss function has the form 
\begin{equation}\small
    \mathcal{L} = \mathbb{E}_{i \sim U\{1,N\}}\left[\mathbb{E}_{q(\tau^i|\tau^{i-1})}\left[\frac{\lambda(\alpha)}{2 \sigma^2_q(t)}\|\tau^{\theta}(\tau^i, t)-\tau^0\|^2_2\right]\right],
\end{equation}
where $\lambda(\alpha) = \frac{\Bar{\alpha}_{i-1}(1-\alpha_i)^2}{(1-\Bar{\alpha}_i)^2}$.

\begin{algorithm}[t]
    \vspace{0.5em}
    \caption{Guided Trajectory Prediction}\label{alg:guide_sampling}
    \KwIn{Observation $X$, intent $G_k$, Gradient Map $\nabla M_d$}
    \KwOut{Trajectory Prediction $\hat{\tau}^0$}
    \vspace{0.5em}
    $\hat{\tau}^N \sim \mathcal{N}(0,\mathbf{I})$\;
    \For{$N = i, \ldots ,0$}{
    \tcp{Denoise Sampling step}\
    $\hat{\tau}^i_{unguided} \sim \mathcal{N}(\mu_{\theta}(\hat{\tau}^{i+1}, i), \sigma^2_q(t)\mathbf{I})$\;
   \tcp{Guidence step}\
   $\hat{\tau}^i = \hat{\tau}^i_{unguided} +  \mathcal{H}(\nabla M_d, \hat{\tau}^i_{unguided})$ \;
    }
    \Return $\hat{\tau}^0$ as the final prediction for $G_k$
\end{algorithm}

\section{Implementation Details}
\label{sec: implementation detail}

The \trajdif model requires the predicted intent of an agent and generates a trajectory prediction based on the predicted intent. Here, we use the CVAE-based Macro-stage of the MUSE-VAE model \cite{lee2022muse} to predict this. The macro stage takes in a heat-map representation of an agent's past trajectory along with the binary scene map; it outputs a distribution for waypoint and goal locations in the form of multichannel location heatmaps, where each channel represents the distribution of possible agent locations for each way-point or goal step. By sampling $K$ latent factors $w$ in the LG-CVAE net, we can decode $K$ sets of way-point and goal pairs to condition on for the \trajdif model. We also point out that, for the following experiments, \trajdif and MUSE-VAE share the same sets of predicted waypoint sets. For the PFSD dataset, we use 25 denoising steps to generate the prediction, and for the nuScenes dataset we use 20 denoising steps. We use the cosine schedule from \cite{nichol2021improved} for the forward noising process. A psudo-code for the sampling denoising process is also provided in \autoref{alg:guide_sampling}.

Following \cite{janner2022diffuser}, our U-net structure that models $P_{\theta}(\tau^{i-1}|\tau^i)$ uses six residual convolution blocks in the encoding and decoding modules, where each residual block contains two 1-D convolution operations along with a group norm operation \cite{Wu_2018_ECCV} and a Mish activation \cite{misra_2019_BMVC}. We train the model with an Adam optimizer with a learning rate of $10^{-3}$ for the PFSD dataset and $2\cdot10^{-4}$ for the nuScenes dataset. For waypoint prediction, we used the pre-trained MUSE-VAE macro stage from the original implementation\footnote{\url{https://github.com/ml1323/musevae}}. The full implementation of \trajdif will be open-sourced upon acceptance.

\section{Metrics}
\label{metric sup}
For the benchmark evaluation, we compute the minimum average displacement error $\mathbf{ADE_k}$ and the final displacement error $\mathbf{FDE_k}$ of the trajectory samples $K$ for each agent compared with the ground truth trajectory: 
\begin{align}
\text{ADE}_k = \frac{1}{T_p} \min_k \sum_{t=T_o+1}^{T_o+T_p} \|y_t-\hat{y_t}^{(k)}\|, \ 
\text{FDE}_k =  \min_k  \|y_{T_o+T_p}-\hat{y}_{T_o+T_p}^{(k)}\|.
\end{align}
Here, $\hat{y}_{t}^{(k)}$ is the predicted coordinate in the $t$th frame for the $k$th sample. We also adopt the Kernel Density Estimated-based Negative Log Likelihood (\textbf{KDE NLL}) proposed in \cite{ivanovic2019trajectron}. This metric uses a kernel density estimate of the sampling distribution probability density function $p_{sample}^t$ for each time step $t$ and calculates the average negative log-likelihood of the ground truth trajectories. We evaluate the environment understanding of the prediction models using the Environmental Collision-Free Likelihood (\textbf{ECFL}) proposed in \cite{sohn2021a2x}. ECFL measures the probability that a predicted trajectory is free of collision with the environment or violates the navigability constraint: 
\begin{align}
\text{ECFL}(\hat{y}^{(k)},M_b) &= \frac{1}{k}\sum_k \prod_{t=T_o+1}^{T_o+T_p} M_b(\hat{y}^{(k)}_t).
\end{align}
The ECFL takes in all the $K$ sampled trajectories and a binary environmental map where "$1$" and "$0$" indicate navigable and non-navigable areas, respectively. The ECFL measures the portion of the $K$ predictions that produce \textbf{no} environmental violations. Finally, we quantify the diversity of the prediction outputs using the Multiverse Entropy \textbf{(MVE)} proposed in \cite{sohn2021a2x}. The MVE quantifies the diversity by looking at the heading of the trajectory predictions.
We calculate the average heading $h^{(k)}$ of a trajectory prediction for the $K$ predictions generated by the models. We then estimate the distribution of the average heading $\hat{p}(H)$ using $K$ outputs from each model, where $H$ is the random variable for the heading, and measure the entropy of the estimated distribution:
\begin{align}
    \text{MVE}(H) = -\sum_{h\in H}\hat{p}(h)\log_2(\hat{p}(h)).
\end{align}
The higher the MVE value, the more diverse the trajectory output.

\section{Benchmark Model Choices}
\label{sec:model choice}

In Sec.4 of the main paper, we benchmark against five SOTA trajectory prediction models. The four non-diffusion-based models are chosen following comparisons in the current literature on trajectory prediction. Among these models, we compared with MUSE-VAE~\cite{lee2022muse} (CVPR 2022). For diffusion-based models, we chose MID~\cite{Gu_2022_CVPR} (CVPR 2022) since the authors have open-sourced their code base for replication and benchmarking on new data sets. Other diffusion-based models reviewed in Sec.2 of the main paper have not open-sourced their code or have not disclosed the source code necessary for model retraining. 

\section{Model Settings}
\label{sec: model setting}
\begin{table}[]
    \centering
    \caption{Comparison of settings. Rows correspond to the context used by the models and columns correspond to the prediction setting of the models.}
    \resizebox{0.75\textwidth}{!}{
    \begin{tabular}{ccc}
        \toprule
        \multirow{2}{*}{Input Context}& \multicolumn{2}{c}{Prediction Configuration}  \\
        \cmidrule(lr){2-3}
          & Independent & Joint\\
         \cmidrule(lr){1-3}
        \multirow{3}{*}{Single-agent} & Y-Net~\cite{mangalam2021goals} &  \\
        & MUSE-VAE~\cite{lee2022muse}  & N/A \\
        & \trajdif & \\
        \cmidrule(lr){1-3}
        \multirow{2}{*}{Multi-agent} & Trajectron++~\cite{ivanovic2019trajectron} & \multirow{2}{*}{AgentFormer~\cite{yuan2021agentformer}} \\
        & MID~\cite{Gu_2022_CVPR} &  \\
        \bottomrule
    \end{tabular} }
    \label{tab:category}
\end{table}

We note that not all models with which we compare are in the same prediction and context settings as our \trajdif. However, the difference in settings does not affect the analysis for the main scope of the study, which is to achieve environmentally compliant trajectory predictions while maintaining the accuracy and diversity of the prediction. 

We categorize the benchmark models into three categories based on their input contexts and prediction settings, as noted in \autoref{tab:category}. \textit{Independent Prediction} models predict the trajectory of an agent independently of the trajectory predictions of other agents in the scene, while \textit{Joint Prediction} models make all agents' predictions simultaneously, dependent on each other.  These predictions can be contextualized on the past trajectories of the single focal agent in \textit{Single-agent context} or all agents in the scene in \textit{Multi-agent context} setting.  \trajdif, MUSE-VAE~\cite{lee2022muse} and Y-Net \cite{mangalam2021goals} belong to the \textit{Independent Prediction / Single-agent Context} setting.  Other works aim to explicitly focus on the social aspect of trajectory prediction; they introduce the multi-agent context but still perform independent agent prediction during the inference phase. Trajectron++ \cite{ivanovic2019trajectron} and MID \cite{Gu_2022_CVPR} belong to this category. Finally, only a few models, notably Agentformer\cite{yuan2021agentformer}, perform joint multi-agent prediction in the multi-agent context.

Differences in Input Context have a minor conceptual modeling impact on the Prediction Configuration setting, adding varied conditioning signals that can be easily carried across different models (e.g., by replacing the context encoder blocks).  This is why our \trajdif comparison with \cite{lee2022muse,ivanovic2019trajectron,Gu_2022_CVPR} remains relevant.  Even when compared to \cite{yuan2021agentformer}, which aims to jointly predict multi-agent trajectories, our \trajdif's independent predictions compare extremely favorably (Tables 1a and 1b in the main paper), despite the theoretical advantage of considering joint predictions to produce socially compliant trajectories. We further discuss this issue below.


\subsection{Socially Compliant Predictions}
\label{sec:ACFL}
To assess the ability of different models to produce socially compliant trajectories, we performed a new benchmark based on the ACFL metric introduced in \cite{sohn2021a2x}.  ACFL evaluates the percentage of prediction outputs that do not involve collisions with all $K$ modes of the other agents' predictions within the same scene. We used different threshold values to determine collision occurrences between two agents in the PFSD and nuScenes datasets. Specifically, we set the thresholds at 0.5 meters for PFSD and 3 meters for nuScenes, selected based on the minimum distance observed between all agent pairs in the ground-truth future trajectories within each dataset.  Note that most prior studies on multi-agent prediction models lack quantitative analysis on how well their models can generate socially compliant trajectory predictions. 

\begin{table}[tbhp]
    \centering

    \caption{Results on PFSD with $K=20$ with $t_{obs}=3.2$s (8 frames) and $t_{pred}=4.8$s (12 frames) with the addition of ACFL metric. Errors are in meters. The best performance is boldfaced and the 2nd place is marked as blue. Numbers in parenthesis indicate the ranking for the score. 
    }
    \resizebox{\columnwidth}{!}{
    \begin{tabular}{l|cccccc}
        \toprule
         Model & ADE $\downarrow$ & FDE $\downarrow$ & NLL $\downarrow$ & ECFL $\uparrow$ & MVE $\uparrow$ & ACFL $\uparrow$\\ \midrule
         Trajectron++ &  0.20 (7)  & 0.42 (7) & 2.24 (7) & 85.00 (7) & \textbf{1.13} (1) & 29.55 (6) \\
         AgentFormer & 0.11  (6)  & 0.17 (5) & \textbf{1.93} (1)& 93.76 (4) & 0.67 (7) & \textbf{53.60 (1)}\\
         Y-net & 0.07 (3) & 0.12 (3) & 1.98 (3) & 94.16 (3) & 0.79 (6) & \textcolor{blue}{49.57 (2)}\\
         MUSE-VAE & \textbf{0.05} (1)& \textbf{0.09} (1) & \textcolor{blue}{1.95} (2) & \textcolor{blue}{97.08} (2) & 0.92 (4) & 41.16 (3)\\
         MID & 0.09 (4) & 0.16 (4) & 2.00 (5) & 88.72 (6) & 0.93 (3) & 28.20 (7)\\
         MID w/Map & 0.10 (5) & 0.19 (6) & 2.00 (5) & 90.41 (5) & 0.86 (5) & 32.88 (4)\\
         \rowcolor{red!20} \textbf{\trajdif} & \textcolor{blue}{0.06} (2)& \textbf{0.09} (1) & 1.98  (3) & \textbf{99.62} (1) & \textcolor{blue}{1.08} (2) & 30.35 (5)\\ \bottomrule
    \end{tabular}
    }
    \label{tab:PFSD_result}
    \end{table}

\begin{table}
    \small
    \centering
    \caption{Result on nuScenes with $K=5$ and $K=10$ with $t_{obs}=2$s (4 frames) and $t_{pred}=6$s (12 frames) with the addition of ACFL metric. Errors are in meters. The best performance is boldfaced and the 2nd place is marked as blue. Numbers in parenthesis indicate the ranking for the score. 
    }
    \resizebox{\columnwidth}{!}{
    \begin{tabular}{c|lcccccc} \hline
         K &  Model & ADE $\downarrow$ & FDE $\downarrow$ & NLL $\downarrow$ & ECFL $\uparrow$ & MVE $\uparrow$ & ACFL $\uparrow$ \\ \hline 
         \multirow{5}{*}{5} &Trajectron++ &  2.51 (7) & 5.57 (6) & 11.66 (7) & 81.66 (4) & 0.46 (6) & 69.94 (6)\\
         &AgentFormer & 1.86 (4) & 3.89 (4) & 6.94 (3) & 84.66 (3) & 0.38 (7) & 65.05 (7)\\
         &Y-net &\textcolor{blue}{1.63} (2) & 2.86 (3) & 7.13 (4) & 76.61 (5) & \textcolor{blue}{0.68} (3) & 77.85 (2) \\
         &MUSE-VAE & \textbf{1.37} (1) & \textcolor{blue}{2.84} (2) & \textbf{5.76} (1) & \textcolor{blue}{89.30} (2) &  0.65 (4) & 77.89 (1) \\
         &MID & 2.38 (5) & 5.54 (5) & 9.33 (5) & 69.23 (6) & \textbf{0.81} (1) & 75.44 (4)\\
         &MID w/Map & 2.42 (6) & 5.61 (6) & 9.51 (6) & 68.72 (7) & \textbf{0.81} (1) & 74.14 (5) \\
         &\cellcolor{red!20}\textbf{\trajdif} & \cellcolor{red!20}1.67 (3) & \cellcolor{red!20}\textbf{2.73} (1) & \cellcolor{red!20}\textcolor{blue}{6.85} (2) & \cellcolor{red!20}\textbf{99.15} (1) & \cellcolor{red!20}0.61 (5) & \cellcolor{red!20}76.08 (3) \\ \hline
         \multirow{5}{*}{10} &Trajectron++ &  1.92 (5) & 4.01 (5) & 8.20 (7) & 81.25 (4) & 0.57 (6) & 64.05 (6) \\
         &AgentFormer & 1.45 (4) & 2.86 (4) & 5.67 (4) & 84.26 (3) & 0.42 (7) & 54.72 (7)\\
         &Y-net & \textcolor{blue}{1.32} (1) & \textcolor{blue}{2.05} (2) & 5.60 (3) & 70.71 (5) & \textbf{1.03} (3) & 70.98 (3)\\
         &MUSE-VAE & \textbf{1.10} (1) & 2.11 (3) & \textbf{4.61} (1) & \textcolor{blue}{89.26} (2) & 0.79 (4) & 74.13 (1)\\
         &MID & 1.93 (6) & 4.29 (7) & 7.42 (6) & 68.97 (6) & \textcolor{blue}{1.00} (2) & 68.61 (4)\\
         &MID w/Map & 1.96 (7) & 4.28 (6) & 7.41 (5) & 68.40 (7) & \textcolor{blue}{1.00} (2) & 68.19 (5)\\
         &\cellcolor{red!20}\textbf{\trajdif} & \cellcolor{red!20}1.41 (3) & \cellcolor{red!20}\textbf{2.02} (1) & \cellcolor{red!20}\textcolor{blue}{5.33} (2) & \cellcolor{red!20}\textbf{99.08} (1) & \cellcolor{red!20}0.74 (5) & \cellcolor{red!20}72.10 (2)\\ \hline
    \end{tabular}
    }
    \label{tab:nuScenes_result}
\end{table}

\paragraph{PFSD.} As shown in \autoref{tab:PFSD_result}, all models struggle to achieve good ACFL scores for the PFSD dataset, a relatively dense multi-agent setting. The PFSD dataset contains on average 14 agents per scene and each pair of agents is 3.8 meters apart on average, with a minimum of 0.5 meters.  AgentFormer, as expected with its joint multi-agent prediction setting, achieves the best performance, yet the model still fails to produce guaranteed socially compliant behavior as half of the trajectory pairs contains collision. Y-Net, with an independent prediction configuration using \textit{single-agent} context surprisingly achieves the second place for ACFL. However, both models struggle to perform in other metrics. Our \trajdif lags in the ACFL metric, as it does not consider other agents during the prediction process, but performs comparably or better than T++ and MID that use a multi-agent context. This demonstrates that the multi-agent input context alone is not sufficient for models to achieve socially feasible predictions.

\paragraph{nuScenes.} \autoref{tab:nuScenes_result} demonstrates that all models perform better in the ACFL metric on the nuScenes dataset relative to their performance on the PFSD dataset. The nuScenes dataset is a sparser multi-agent setting, with 4 agents per scene, each pair of agents being 42 meters apart on average with a minimum of 5 meters. For both $K=5$ and $K=10$ settings, independent agent prediction models (Y-Net, MUSE-VAE, and \trajdif) have the best ACFL performance. Our \trajdif ranked 2 and 3 in the settings $K=5$ and $K=10$, respectively, following very closely the top performing MUSE-VAE. All models using the multi-agent context performed worse. AgentFormer, the best ACFL performer on PFSD, has the worst ACFL for nuScenes. 

With the addition of the ACFL metric, we wish to present a fair and comprehensive comparison between models across different settings. Our \trajdif is able to match SOTA level ACFL performance for the sparser nuScenes dataset where multi-agent interactions are not as prevalent. For the denser PFSD dataset, our model underperforms as it neglects information about other agents during the prediction process. However, our model still matches or outperforms the ACFL performance of models performing independent prediction with multi-agent contexts and outperforms AgentFormer in all other accuracy, diversity, and ECFL metrics. We argue that although modeling the social aspect of human trajectory prediction is an important problem, our model was able to advance environmental understanding while often matching the SOTA ACFL scores and we left achieving socially compliant multi-agent HTP model as an exicting future direction. 

\section{Additional Qualitative Evaluations}
\label{sec:additional result}
\begin{figure*}
    \centering
    \includegraphics[width=1\textwidth, height=1\textwidth]{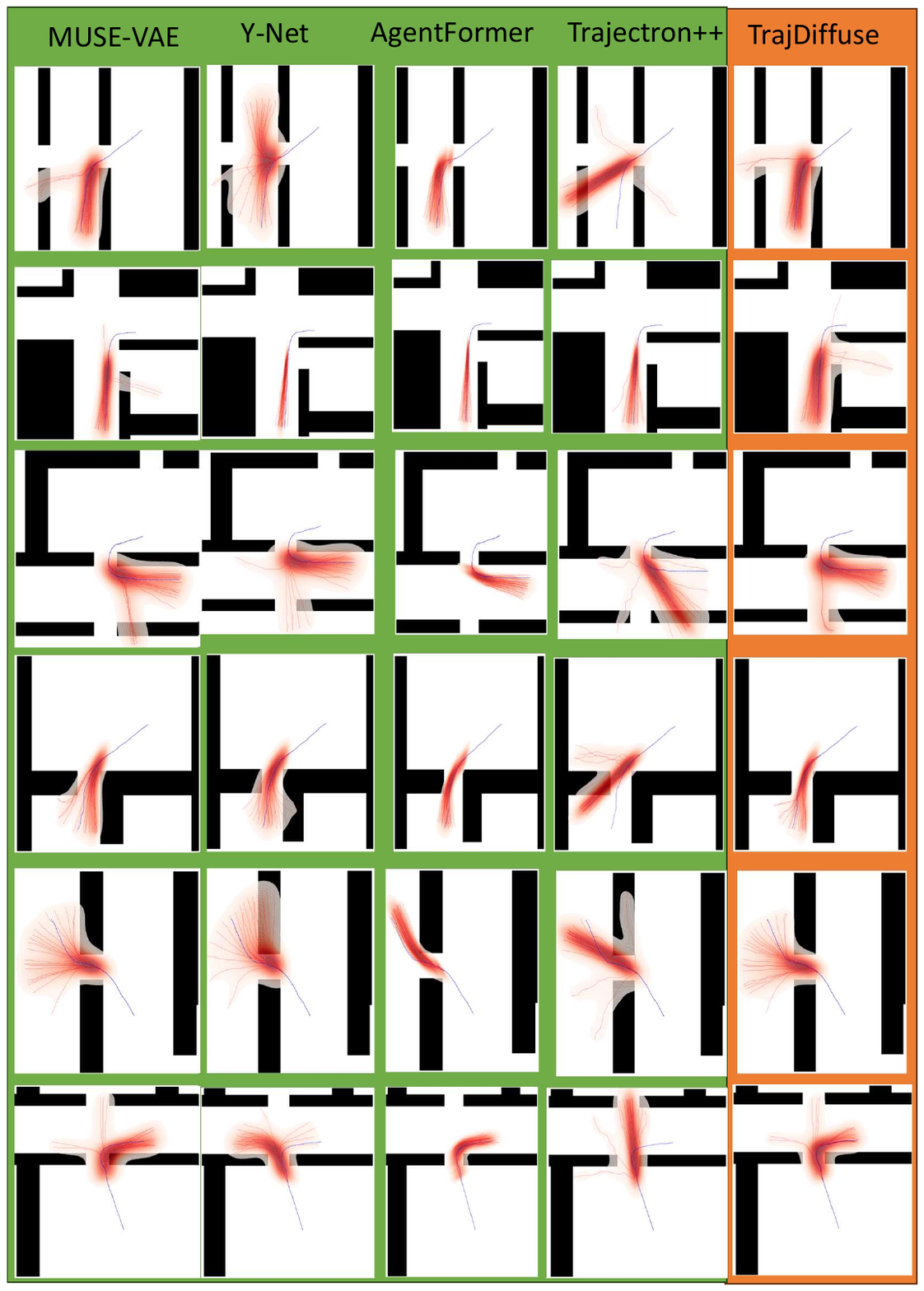}
     \caption{Visualizations for PFSD with complex layouts and hard maneuvers with $K= 20$. Each column contains visualizations of an agent's trajectories predicted by the model indicated at the top of the column. Each row corresponds to the agent with identical initial conditions and identical prior motion history. The blue dashed line indicates the observed and GT trajectory; the red dashed line indicates the predicted trajectory. }
     \label{fig:PFSD_supplement}
     \vspace{-0.1in}
\end{figure*}

\begin{figure*}
    \centering
    \includegraphics[width=1\linewidth]{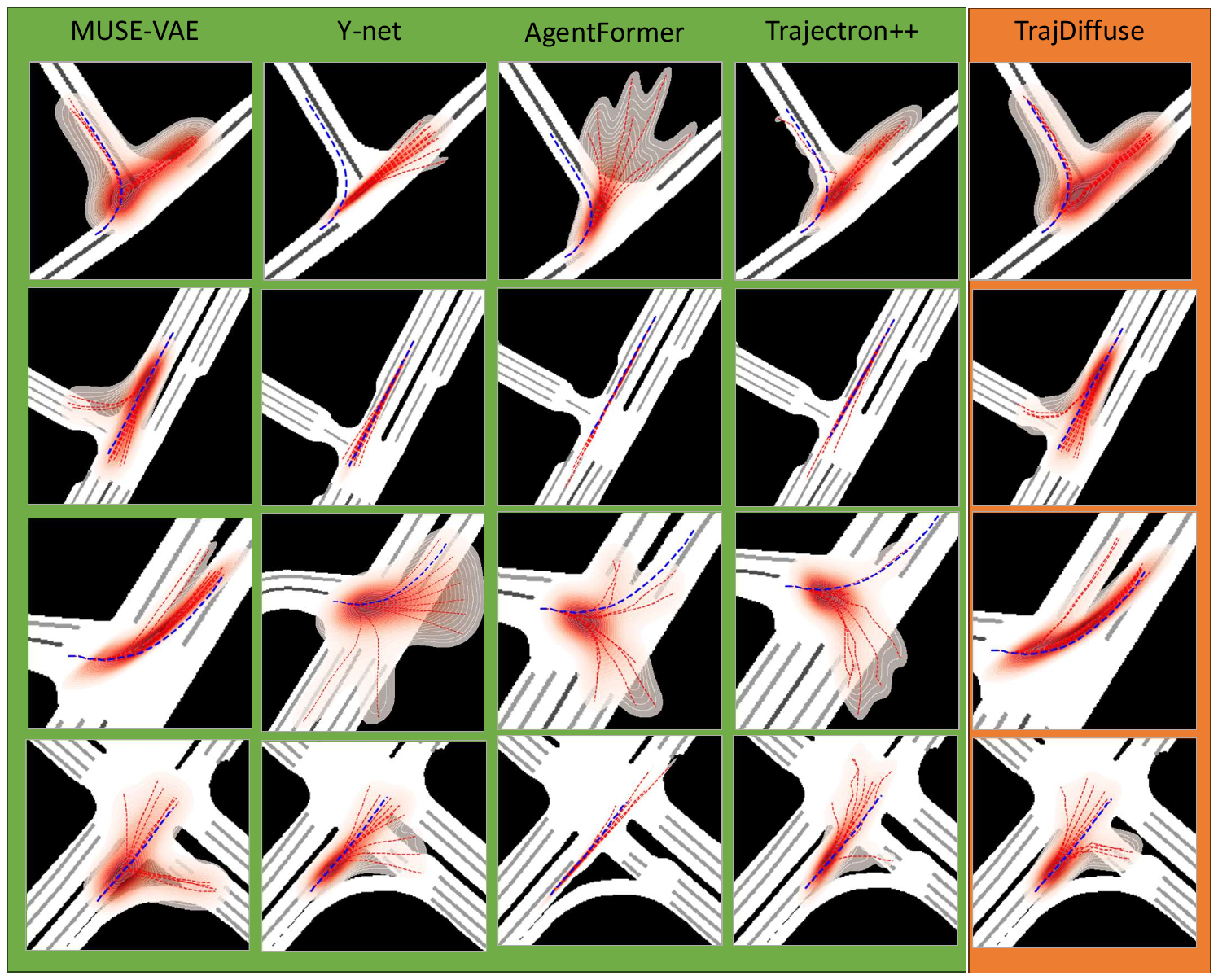}
     \caption{Visualizations for hard nuScenes instances with slow observed trajectories with $K= 10$. Each column contains visualizations of an agent's trajectories predicted by the model indicated at the top of the column. Each row corresponds to the agent with identical initial conditions and identical prior motion history. The blue dashed line indicates the observed and GT trajectory; the red dashed line indicates the predicted trajectory.}
     \label{fig:nuScenes_supplement}
     \vspace{-0.1in}
\end{figure*}

We offer visualizations in \autoref{fig:PFSD_supplement} and \autoref{fig:nuScenes_supplement} of additional samples from the benchmark datasets, which showcase the characteristics of the models as a supplement to our experiments on the PFSD and nuScenes in the main article. In \autoref{tab:PFSD_result_supp} and \autoref{tab:nuScenes_result_supp}, we provide quantitative results for the instances shown in the visualizations.

\paragraph{PFSD.} \autoref{fig:PFSD_supplement} contains visualizations of new scenarios from the PFSD dataset. Here, the instances are mostly non-linear maneuvers of agents taking sharp turns or consecutive turns around obstacles. Each environment layout contains multiple possible destinations following the turn. Therefore, these represent challenging real-world scenarios for assessing the ability of models to predict multiple diverse outcomes. Our \trajdif is capable of excelling in these more complex scenarios by producing predictions that are accurate, feasible, and diverse. Our model is able to replicate the agent's original maneuver and reach the ground truth destination; it is also capable of generating other possible trajectories that strictly follow the environment layout with the help of the map-based guidance term. Agentformer can mimic the ground-truth trajectory; however, its output lacks diversity. The model output always concentrates around paths toward the same heading directions and fails to explore other feasible destinations. This also aligns with the quantitative result of \autoref{tab:PFSD_result_supp}, where Agentformer has the lowest MVE score. Trajectron++ is capable of predicting more diverse trajectory outputs than Agentformer; however, it often fails to generate accurate predictions. This model cannot anticipate the accurate intention of the agent and fails to predict the correct maneuvers. This is also reflected in the poor ADE and FDE measurements in \autoref{tab:PFSD_result_supp}. The two MID models also struggle to predict the correct movement in these harder cases, as both models have poor ADE and FDE performance for these instances. The base model also struggles to understand environmental constraints as it often generates predictions ignoring the environment layout. The addition of map embedding only offers a small improvement but does not fundamentally change the model behavior. Y-net is able to achieve the most diverse predictions with the help of the test-time sampling trick (TTST). However, the output of the model often violates environmental constraints. This again demonstrates that TTST causes a trade-off between diversity and the feasibility of the prediction output. Y-net also has one of the lowest ECFL among the seven models in these instances. The MUSE-VAE model output is more balanced between accuracy, diversity, and feasibility with the help of the two-stage planning strategy, the heat map-based representation, and the CVAE-based model design. However, its RNN-based micro-stage neither takes in any environmental information nor uses the environment-aware macro-stage output directly as waypoints. Therefore, its output struggles in these more complex maneuvers.  MUSE-VAE's output often violates environmental constraints while trying to reach the predicted goal, causing unrealistic predictions and sub-par numerical performance as shown in \autoref{tab:PFSD_result_supp}.

\paragraph{nuScenes.} In \autoref{fig:nuScenes_supplement}, we provide additional visualizations from the experiment with the $K=10$ configuration. These instances involve observed slow-velocity trajectories in complex and narrow layouts with multiple possible destinations. This makes the prediction of the true intention of the agent even more difficult. The nature of the nuScenes dataset as a vehicle-focused trajectory dataset also requires the model to have a better understanding of the navigability in the environment.  Our \trajdif is capable of achieving good prediction accuracy while maintaining diversity and feasibility in these difficult cases. It does not produce off-road trajectory predictions and explores other possible destinations in addition to ground-truth trajectories. From \autoref{tab:nuScenes_result_supp}, we can see that in these harder cases \trajdif is able to achieve the best FDE and the second place in the ADE. It is also capable of achieving perfect ECFL performance despite the narrower road layout and a good MVE score. The sequence-to-sequence-based Trajectron++ and Agentformer struggle to predict the correct intention of the agents given the slow observed trajectories. Trajectron++ often fails to accurately reach the destination of the agent. It is capable of achieving some diversity in its prediction, but these outputs often end up in off-road areas. Agenformer also struggles to generate accurate predictions and has poor environmental understanding. The MID and its variant also struggle to reach the intended goal of the agent, which is also reflected in the poor FDE performance for these instances. The MID model also often ignores map constraints; the addition of map embedding also does not provide enough context for the model to produce more environmentally feasible predictions. The Y-net model achieved the best MVE performance in these instances. However, it also has one of the worst ECFL performance among the benchmark models in these more environmentally restricted cases, again demonstrating the trade-off between diversity and feasibility due to the non-data-driven sampling trick. The MUSE-VAE model is able to achieve good performance on accuracy measures; however, its micro-stage often produces cases that violate the environmental constraints in these narrower road settings.

\section{Inference Speed}
\label{sec:inference speed}
We perform evaluation of the inference time for MUSE-VAE (3.69s), MID (6.89s), and our \trajdif (1.59s) on a PFSD scene with 8 agents generating 20 samples each. \trajdif required the least time, using only 25 diffusion steps due to our diffusion-as-interpolation approach, compared to 100 steps for MID (6.89s). Although \trajdif achieves better inference speed among the benchmarked model here, enhancing HTP inference speed remains a critical area for future research. \trajdif inference could also be further benefit from alternative diffusion sampling scheme such as DDIM \cite{song2020denoising} or consistency model \cite{song2023consistency}. We note that this speed comparison might not be the most rigorous reference for downstream deployment or applications and computational complexity is also not the main focus of the paper. We wish to showcase through this comparison that our diffusion formulation is superior compare with previous attempt (MID) and can perform favorable compare with other HTP models (MUSE-VAE) as well.
\section{Limitations and Challenges}
\label{sec: limitation}
\begin{figure*}
    \centering
    \includegraphics[width=1\linewidth]{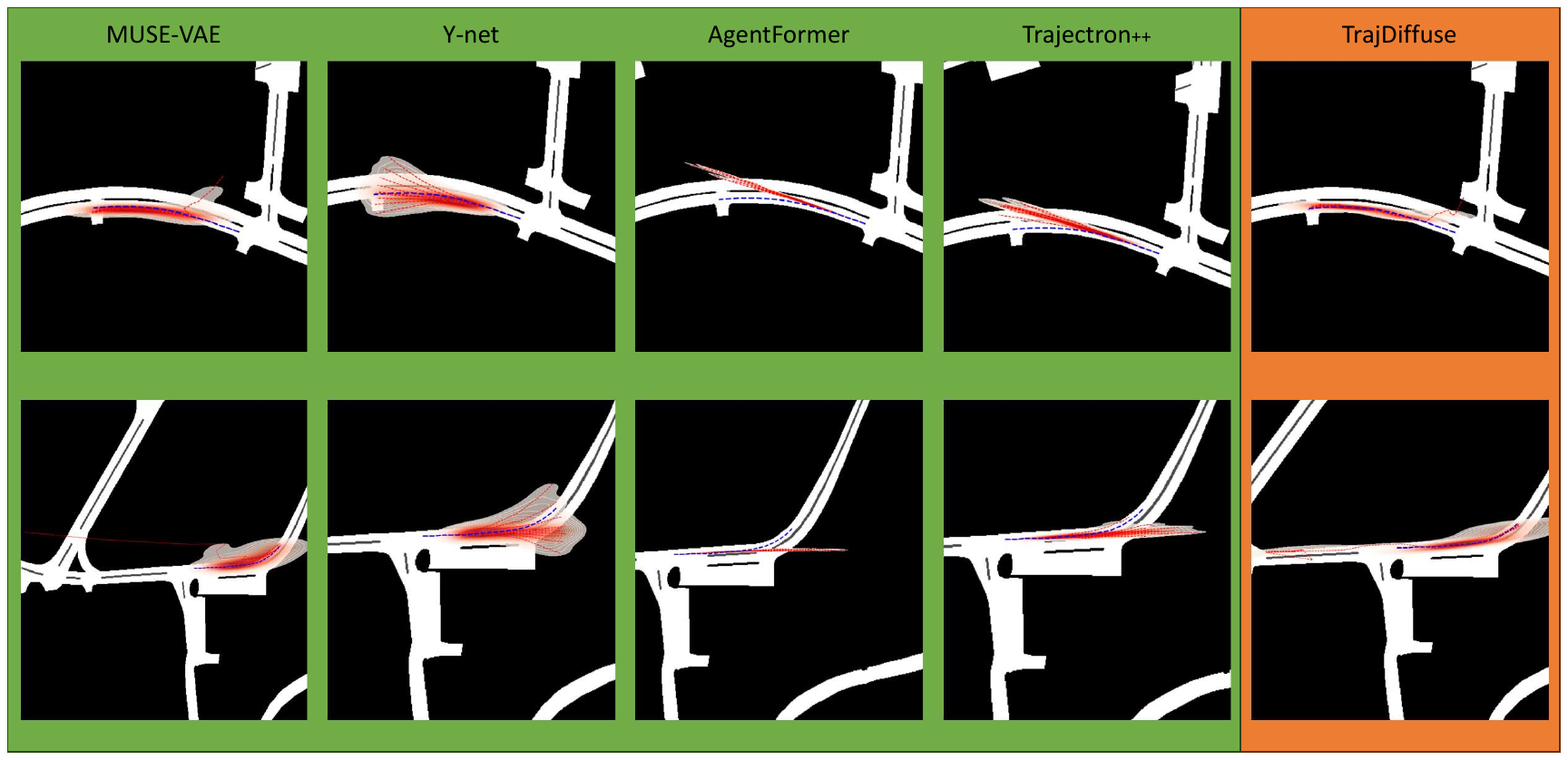}
     \caption{Visualizations for nuScenes with $K= 10$: Two Challenging Cases for \trajdif. Each column contains visualizations of an agent's trajectories predicted by the model indicated at the top of the column. Each row corresponds to the agent with identical initial conditions and identical prior motion history. The blue dashed line indicates the observed and GT trajectory; the red dashed line indicates the predicted trajectory.}
     \label{fig:failed_nuScenes_supplement}
     \vspace{-0.1in}
\end{figure*}

In our experiments, as noted in Sec.5 of the main article, we used MUSE-VAE's macro-stage to predict the waypoints conditioning input for our \trajdif. In certain cases, the micro-stage might produce waypoint predictions that are off by a significant margin; these cases would make it challenging for the \trajdif to generate a reasonable trajectory prediction, and hence the ECFL performance in Sec.4 is not perfect at 100\%. Here, we present two of the examples from the nuScenes dataset where \trajdif fails in \autoref{fig:failed_nuScenes_supplement}. 

In the first case, the \trajdif is conditioned on a set of poorly predicted waypoints. These waypoints are off by a great margin and cause our model to violate the navigable area constraint. We see in the MUSE-VAE output that the sample that used the same waypoint set also experienced a similar issue. For the second case, one of the \trajdif outputs is conditioned on a set of unrealistic waypoints, which causes the trajectory output to go in the opposite direction. Despite traveling in the wrong direction, map-based guidance is able to keep the trajectory inside the drivable area. However, for the MUSE-VAE model, the trajectory generated based on the same set of waypoints goes off-road and violates the navigability constraint. 

Although our model is able to adhere to environmental constraints for most cases, a failure might occur when the waypoint prediction module substantially fails. A dedicated waypoint prediction module that is better end-to-end integrated into the \trajdif's conditional generation pipeline might relieve this issue, and we leave it as an open research challenge for future study.

\begin{table}
    \centering
    \caption{Results on PFSD for instances in \autoref{fig:PFSD_supplement} with $K=20$ with $t_{obs}=3.2$s (8 frames) and $t_{pred}=4.8$s (12 frames). Errors are in meters. The best performance is boldfaced and the runner-up is marked as blue.}
  \resizebox{\columnwidth}{!}{
  \begin{tabular}{lccccc} \toprule
         Model & ADE $\downarrow$ & FDE $\downarrow$ & KDE NLL $\downarrow$ & ECFL $\uparrow$ & MVE $\uparrow$\\ \midrule
         Trajectron++ &  0.82& 1.78 & 3.36 & 52.67 & 1.28 \\
         AgentFormer & \textcolor{blue}{0.12} & \textcolor{blue}{0.16} &\textbf{1.94} & \textcolor{blue}{98.33} & 0.71 \\
         Y-net & 0.14 & 0.20 & 2.38 & 82.5 & \textbf{1.75} \\
         MID & 0.39 & 0.58 & 2.38 & 46.67 & 1.59 \\
         MID (w/ map) & 0.20 & 0.28 & 2.26 & 58.33 & 1.64 \\
         MUSE-VAE & 0.25 & 0.30 & 2.18 & 87.5 & 1.51 \\
         \rowcolor{red!20}\textbf{\trajdif} & \textbf{0.09} & \textbf{0.15} & \textcolor{blue}{2.14}  & \textbf{100.0} & \textcolor{blue}{1.58} \\ \bottomrule
    \end{tabular}
    }
    \label{tab:PFSD_result_supp}
\end{table}

\begin{table}
    \small
    \centering
    \caption{Result on nuScenes for instances in \autoref{fig:nuScenes_supplement} with $K=10$ with $t_{obs}=2$s (4 frames) and $t_{pred}=6$s (12 frames). Errors are in meters. The best performance is boldfaced and the runner-up is marked as blue.}
  \resizebox{\columnwidth}{!}{
    \begin{tabular}{lccccc}
        \hline
         Model & ADE $\downarrow$ & FDE $\downarrow$ & KDE NLL $\downarrow$ & ECFL $\uparrow$ & MVE $\uparrow$\\ \hline 
         Trajectron++ & 2.37 & 5.27 & 8.61 & \textcolor{blue}{70.0} & \textcolor{blue}{1.06} \\
         AgentFormer & 2.80 & 5.76 & 11.66 & \textcolor{blue}{70.0} & 0.93 \\
         Y-net & 4.40 & 8.53 & 36.50 & 65.0 & \textbf{1.12} \\
         MUSE-VAE & \textbf{0.93} & \textcolor{blue}{1.72} & \textbf{3.38} & 57.5 &  0.95 \\
         MID & 2.79 & 6.64 & 11.71 & 62.5 & 1.00 \\
         MID (w/ map) & 2.42 & 5.63 & 9.75 & 60.00 & 0.88 \\
         \rowcolor{red!20}\textbf{\trajdif} & \textcolor{blue}{1.55} & \textbf{0.44} & \textcolor{blue}{4.48} & \textbf{100.0} & 1.00\\ \hline
    \end{tabular}
    }
    \label{tab:nuScenes_result_supp}
\end{table}

\end{document}